\documentclass[10pt,twocolumn,letterpaper]{article}

\usepackage{cvpr}
\usepackage{times}
\usepackage{epsfig}
\usepackage{graphicx}
\usepackage{amsmath}
\usepackage{amssymb}

\usepackage[breaklinks=true,bookmarks=false]{hyperref}

\cvprfinalcopy 

\def\cvprPaperID{****} 

\usepackage{booktabs}       
\usepackage{nicefrac}       
\usepackage{microtype}      

\usepackage{thmtools}
\declaretheorem{theorem}
\declaretheorem{corollary}

\declaretheorem{definition}
\declaretheorem{remark}

\usepackage{subfig}
\usepackage{multirow}
\usepackage{float}
\usepackage{rotating} 
\usepackage{adjustbox}

\usepackage{tabulary}

\everypar{\looseness=-1 } 
\newcommand{\textcite}[1]{\cite{#1}}

\usepackage{wrapfig}

\usepackage{tikz}
\usetikzlibrary{decorations.text,calc,shapes,arrows,arrows.meta, positioning,shapes.misc,decorations.markings,decorations.markings,decorations.pathreplacing,matrix,spy}
\usepackage{pgfplotstable}
\usepackage{units,pgfplots}
\usepgfplotslibrary{groupplots}

\makeatletter
\pgfplotsset{
    every axis x label/.append style={
        alias=current axis xlabel
    },
    legend pos/outer south/.style={
        /pgfplots/legend style={
            at={%
                (%
                \@ifundefined{pgf@sh@ns@current axis xlabel}%
                {xticklabel cs:0.5}%
                {current axis xlabel.south}%
                )%
            },
            anchor=north
        }
    }
}
\makeatother

\begin{document}

\title{Robust Design of Deep Neural Networks against Adversarial Attacks based on Lyapunov Theory}
%
 \author{
 	Arash Rahnama\\
 	Booz Allen Hamilton \\ 
  \texttt{Rahnama\_Arash@bah.com} 
  \and
  Andre T. Nguyen \\
  Booz Allen Hamilton \\
  \texttt{Nguyen\_Andre@bah.com} 
  \and
  Edward Raff \\
  Booz Allen Hamilton \\
  \texttt{Raff\_Edward@bah.com} 
 }
 


\maketitle
\begin{abstract}
	Deep neural networks (DNNs) are vulnerable to subtle adversarial perturbations applied to the input. These adversarial perturbations, though imperceptible, can easily mislead the DNN. In this work, we take a control theoretic approach to the problem of robustness in DNNs. We treat each individual layer of the DNN as a nonlinear dynamical system and use Lyapunov theory to prove stability and robustness locally. We then proceed to prove stability and robustness globally for the entire DNN. We develop empirically tight bounds on the response of the output layer, or any hidden layer, to adversarial perturbations added to the input, or the input of hidden layers. Recent works have proposed spectral norm regularization as a solution for improving robustness against $\ell_2$ adversarial attacks. Our results give new insights into how spectral norm regularization can mitigate the adversarial effects. Finally, we evaluate the power of our approach on a variety of data sets and network architectures and against some of the well-known adversarial attacks. 
\end{abstract}
\section{Introduction} \label{sec:intro}
The objective of a supervised learning task for the input $u \in \mathbb{R}^d$, and its associated target value $y$ for a given DNN denoted by $H_\theta(u)$, where $\theta$ is a set of parameters to be learned during the training, is to classify the instance $u$ correctly such that $y=H_\theta(u)$. Recently, the research community has become interested in adversarial attacks, where the adversary's goal is to introduce a small amount of engineered perturbation $\Delta \in \mathbb{R}^d$ to $u$, so that $u^\prime = u + \Delta$, while still maintaining its similarity to $u$, can deceive the DNN into making a mistake, i.e., $H_\theta(u + \Delta) \neq y$. The adversary is usually assumed to be constrained by an $\ell_p$-norm so that $\|u^\prime-u\|_{\ell_p} \leq \epsilon$, where $\epsilon$ bounds the adversaries' freedom to alter the input. While this does not capture the full scope of potential adversaries \cite{Brown2018}, attacks of this form have proven difficult to prevent \cite{szegedy2013intriguing,Carlini:2017:AEE:3128572.3140444,Athalye2018,Su2017}. While optimal defenses have been developed for simple linear models \cite{Biggio2010,Liu:2017:RLR:3128572.3140447}, the over-parameterized nature of DNNs and the complexity of surfaces learned during training make the development of robust solutions against the adversary difficult \cite{Gilmer2018}.

In this work, we use Lyapunov theory of stability and robustness of nonlinear systems to develop a new understanding of how DNNs respond to changes in their inputs, and thus, to adversarial attacks under the $\ell_2$ norm framework. By treating each layer $l$ in the DNN as a nonlinear system $h_l$, we develop a new framework which sets tight bounds on the response of the individual layer to adversarial perturbations (maximum changes in $\| h_l(u) - h_l(u + \Delta) \|_2^2$) based on the spectral norm of the weights, $\rho(W_l)$, of the individual layer. We characterize Lyapunov properties of an individual layer and their relationship to local and global stability and robustness of the network. Since our analysis is based on a sequence of nonlinear transformations $h_{l=1, \ldots, n}$, our method is the first to bound the response to input alterations by the attack parameter $\Delta_{l}$ for any layer $l$ in the DNN. For simple forward networks (fully connected and convolutional), our results show that an attack's success is independent of the depth of the DNN and that the spectral norm of the weight matrix for the first and last layer of a network have the largest effect on robustness. Our Lyapunov analysis of Residual Blocks shows that the skip-connections of these blocks contribute to their lack of robustness against adversarial attacks \cite{Geirhos2019} and that these blocks require more restrictive Lyapunov conditions (a tighter spectral norm regularization) for maintaining the same level of robustness. This may compromise the DNN's accuracy on the clean data set \cite{tsipras2018robustness,couellan2019coupling}. Finally, our proposed robust training method, unlike the previous works in this arena \cite{farnia2018generalizable,qian2018l2,yoshida2017spectral,cisse2017parseval}, regularizes the spectral norm of the weight matrix at each individual layer based on certain Lyapunov conditions, independently from other layers. Our layer-wise spectral regularization parameters may be selected based on our Lyapunov conditions and the level of robustness and accuracy required for a specific task. Our approach can help with the limitations associated with expressiveness of DNNs trained with the same level of spectral regularization across all layers \cite{couellan2019coupling}. We show that this higher degree of freedom in selecting the hyper-parameters for each layer independently, leads to networks that are more accurate and robust against adversarial attacks in comparison to the existing works in the literature. 

In summary, our contributions revolve around showing that adversarial ML research can leverage control theory to understand and build defenses against strong adversaries. This can accelerate the research progress in this area. We prove that with our proposed training approach, the perturbation to the final activation function ($\Delta_n \in \mathbb{R}^d$) is bounded by $\|\Delta_n\|_2 \leq \sqrt{c} \cdot \epsilon$, where $c$ is a constant determined by the hyper-parameters chosen and $\epsilon$ models the adversarial perturbations. Our bound is tight, and applies to \textit{all} possible inputs, and to attacks applied at \textit{any} layer of the network (input, or hidden), with no distributional assumptions. Our analysis shows how Residual Blocks can aid adversarial attacks, and extensive empirical tests show that our approach's defensive advantages increase with the adversary's freedom $\epsilon$. 
\section{Related work} \label{sec:rel}
To provide certifiable defenses, complex optimization schemes are usually adopted to show that all the data points inside an $\ell_p$ ball around a sample data point have the same prediction \cite{wong2018provable,katz2017reluplex,cheng2017maximum,cohen2019certified}. The bounds provided by these methods are usually loose, and the computational costs associated with them increase exponentially with the size of the network. These approaches are only applicable to parts of the input space for which feasible solutions exist. Works such as \cite{zantedeschi2017efficient} have empirically shown that bounding a layer's response to the input generally improves robustness. Works such as \cite{weng2018towards,zhang2018efficient} focus on certifying robustness for a DNN by calculating the bounds for the activation functions' responses to the inputs. The closest to our work are the results given in \cite{farnia2018generalizable,qian2018l2,yoshida2017spectral,cisse2017parseval}. \cite{cisse2017parseval} utilizes Lipschitz properties of the DNN to improve robustness against adversarial attacks. Unlike \cite{cisse2017parseval}, our approach does not require a predetermined set of hyper-parameters to prove robustness. Our analysis provides a range of possible values which determine different levels of robustness and may be selected per application. Similarly, \cite{qian2018l2} empirically explores the benefits of bounding the response of the DNN by regularizing the spectral norm of layers based on the Lipschitz properties of the DNN. These Lipschitz based approaches may be seen as one subset of our Lyapunov based approach. As we will describe, our Lyapunov based analysis is built upon a more general input-output nonlinear mapping which does not necessarily depend on restricting the networks's Lipschitz property. \cite{yoshida2017spectral} explores the benefits of training networks with spectral regularization for improving generalizability against input perturbations. However, their work bounds the spectral norm of all layers by 1. As we will show, this approach limits the performance on the clean data set and does not produce the most robust DNN. \cite{farnia2018generalizable} uses PAC-Bayes generalization analysis to estimate the robustness of DNNs trained by spectral regularization against adversarial attacks. The analysis given in their work however, requires the same regularization condition enforced across all layers of the DNN. Our work provides per layer conditions for robustness, which may be utilized for the selection of the best regularization parameters for each layer independently, given a data set and architecture through cross-validation. Although, we have not empirically shown in the paper, our approach theoretically should provide a level of robustness against intermediate level attacks recently introduced in \cite{huang2018intermediate}. Our work provides a theoretical backing for the empirical findings in \cite{qian2018l2,wong2018provable} which state that Leaky ReLu, a modified version of the ReLu function, may be more robust comparatively. Finally, only one prior work has looked at control theory for adversarial defense, but their method covered only a toy adversary constrained to perturbing the input by a constant \cite{Rahnama2019}. 
\section{Preliminaries and Motivation} 
\label{sec:pre}
\begin{figure}%
	\centering
	\subfloat{{\includegraphics[width=3.8cm]{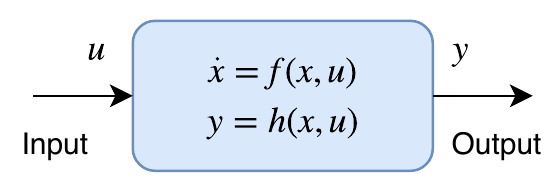} }} \hspace{0.1\textwidth}%
	\subfloat{{\includegraphics[width=6.8cm]{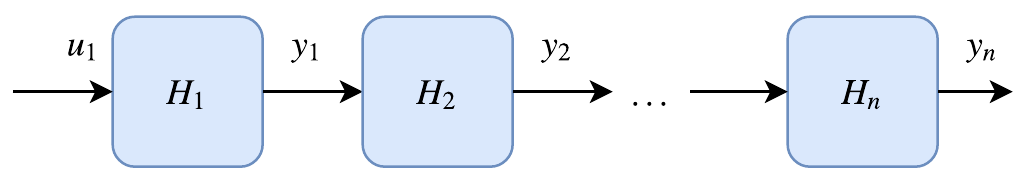} }}%
	\caption{A nonlinear system (top). The DNN modeled as a cascade of nonlinear systems (bottom).}%
	\label{fig:dynsys_cascade}%
\end{figure}

In this work, we address the machine learning problem of adversarial attacks on DNNs from a control theoretic perspective. To aid in bridging the gap between these two fields, we will briefly review the core control theory results needed to understand our work, with further exposition in Appendix \ref{controltheory} for less familiar readers. The design of stable and robust systems that maintain their desired performance in the presence of external noise and disturbance has been studied in the control theory research literature. Our work is based on the Lyapunov theory of stability and robustness of nonlinear systems which dates back to more than a century ago \cite{khalil2002nonlinear}. We treat each layer of the DNN as a nonlinear system and model the DNN as a cascade connection of nonlinear systems. A nonlinear system is defined as a system which produces an output signal for a given input signal through a nonlinear relationship. More specifically, consider the following general definition for the nonlinear system $H$ (Fig. \ref{fig:dynsys_cascade}), 
\begin{equation*} \label{eq:dynsys}
H:
\begin{cases}
\dot{x}= f(x,u) 
&\\y= h(x,u),
\end{cases}
\end{equation*}
where  $x \in X \subseteq 
R^n$, $u \in U \subseteq
R^m$, and $y \in Y
\subseteq R^{k}$ are respectively the state, input and output of the system, and $X$, $U$ and $Y$ are the local state, input and output sub-spaces around the current operating points. The nonlinear mappings $f$ and $h$ model the relationship among the input signal $u$, the internal states of the system $x$ and the output signal $y$. To help connect these control theory notations and definitions to our analysis of DNN models, see Table \ref{tab:table_notation}.  

\begin{table}[!h]
\centering
\caption{A control theory to machine learning mapping.}
\label{tab:table_notation}
\begin{tabulary}{\columnwidth}{|C|L|L|}
\hline
            & Control Theory Meaning                                                                                                                   & Context for our work                                                                                                                                     \\ \hline
$u$         & Input of the nonlinear system                                                                                                            & Inputs to the DNN (e.g., image) or the inputs to a hidden layer from the previous layer                \\ \hline
$y$         & Output of the nonlinear system                                                                                                           & Output of the DNN or the output of any hidden layer                                                                                                      \\ \hline
$x$         & States of the nonlinear system                                                                                                           & Weights and biases of a layer                                                                                                                            \\ \hline
$\dot{x}$   & Transient changes of the states over discrete or continuous steps                                                                        & Changes of the weights and biases during $t$ training steps                                                                                              \\ \hline
$f(.)$      & Nonlinear function modeling the transient changes of the states of the system based on the previous state and current input              & Models the updates applied to the weights' and biases' of a layer over the $t$ training steps                                                            \\ \hline
$h(.)$      & Nonlinear function modeling the steady-state behavior of the system given the current state and input                                    & Models the input-output relationship of a hidden layer given the current values of weights, biases and input                                             \\ \hline
$\rho, \nu$ & Lyapunov parameters modeling the input-output behavior (robustness and stability properties) of the system (explained in Subsection 4.3) & Values determining the extent of spectral regularization enforced at a layer given the desired level of robustness and accuracy (Theorem 3, Corollary 2) \\ \hline
\end{tabulary}
\end{table}

The behavior of a layer inside the DNN may be modeled as a nonlinear system defined above. More specifically, for the layer $l$, the input signal $u$ takes the size of the layer $l-1$ and stands for the input to the layer $l$ before it is transformed by the weights and biases. $y$ has the size of layer $l$ and may be defined as the output of the layer $l$ after the activation functions. The weights and biases of the DNN are the states of the nonlinear systems. In this vein, $h$ and $f$ are general functions which model the relationship between the states $x$, and the nonlinear transformation applied to the input of the layer $u$ to produce the output $y$ of the layer after activation functions. The states are dynamically updated through gradient descent, given the inputs from the training data set during the training iterations $t$. $\dot{x}_l$ is the derivative taken over training iterations indicating that the weights and biases are changing during training, and $f$ models this nonlinear behavior during the training iterations. Please see Appendix \ref{controltheory} for further details.

Our analysis is based on Lyapunov theory which gives us the freedom to define stability and robustness purely based on the input-output relationship of the layers without the exact knowledge of the internal state changes $\dot{x}$ (i.e., we do not need to know the specific weight and bias values). More specifically, Lyapunov theory combined with the fact that we are showing bounded-input-bounded-output (BIBO) stability and robustness of the layer, allows us to abstract out the transient behavior of the nonlinear systems during the training iterations $t$, i.e., $\dot{x}_l(t)=f_l(x_l(t),u_l(t))$ where $x_l(t)=\{W_l(t),B_l(t)\}$, from the robustness analysis and focus only on the steady-state behavior of the nonlinear system, i.e., the input-output mapping of the layer, $y_l=h_l(x_l, u_l)=h_l(\{W_l, B_l\}, u_l)$ where $h_l(.)$ models the nonlinear transformation. By analyzing $h_l(.)$, we can derive robust conditions to be enforced during training for the states of a layer $\{W_l, B_l\}$.

Next, we define an input-output stability and robustness Lyapunov criterion for a nonlinear system. A nonlinear system is said to be BIBO stable, if it produces bounded outputs for bounded input signals. A nonlinear system is said to be stable and robust, if it produces bounded output signals that are close in the Euclidean space, when their respective input signals are also close enough in the Euclidean space \cite{khalil2002nonlinear}. Mathematically, these definitions can be represented as follows,
\begin{definition} \cite{khalil2002nonlinear} \label{def:FG}
	System $H$ is instantaneously incrementally finite-gain (IIFG) $\ell_2$-stable and robust, if for any two inputs $ u_1, u_2\in U$, there exists a positive gain $ \gamma$, such that the relation,
	\begin{align*}
	&||y_{2}-y_{1}||_{\ell_2} \leq \gamma||u_{2}-u_{1}||_{\ell_2}.
	\end{align*}
	holds. Here, $||y_{2}-y_{1}||_{\ell_2}$ and $||u_{2}-u_{1}||_{\ell_2}$ represent the Frobenius $\ell_2$-norm of the input signals $u_1$ and $u_2$ and their respective output signals $y_1$ and $y_2$.
\end{definition}
If a system is IIFG stable and robust, then the changes in the output of the entire system are bounded by the changes in the input of the entire system. As a result, if the changes in the input are minuscule, which is the assumption for the majority of $\ell_2$ adversarial attacks, then the changes in their respective outputs are also minuscule. This is the exact behavior that we would like to encourage for each layer of the DNN so that the entire network is robust against adversarial attacks.
\begin{remark}
	It is important to note that the relationship given in Definition \ref{def:FG} is more general than enforcing Lipschitz continuity. In particular, the above relationship should only hold locally for the input signals $u_1$ and $u_2$ for the DNN to be IIFG. Further, the above assumption does not place any constraints on the initial conditions of the DNN. Additionally, Lipschitz continuity implies uniform continuity, but the relationship given above potentially allows for discontinuous distributions and does not enforce any continuously differentiable condition on the mapping from the input to the output of the DNN \cite{khalil2002nonlinear,fawzi2018adversarial,jin2018stability}. Finally, we will show that the enforcement of the relationship given in Definition \ref{def:FG} locally at the layer level is not necessary. We will show that by enforcing a much looser condition at each layer, we can encourage the behavior given in Definition \ref{def:FG} globally for the DNN.
\end{remark}		
Lyapunov theory of dissipativity  provides a fundamental framework for the stability and robustness analysis of systems based on a generalized notion of the "energy" supplied and dissipated in the system \cite{khalil2002nonlinear}. This generalized notion of "energy" is defined based on a relationship that solely relies on the input and output of the system. This theory states that a passive system which dissipates energy is robust and stable. The benefit of this approach is that by only enforcing a relationship between the input and output of the DNN, we can define a measure of stability and robustness for the entire DNN against adversarial attacks and characterize robust conditions for the states of the DNN (weights and biases). This is done by first defining an input-output mapping which characterizes a specific relationship between the input and output of the layers inside the DNN and then enforcing that the relationship should hold for the DNN for all the inputs supplied to the model and the outputs produced by the model during training so that the same will hold during inference. The following definition provides the mathematical representation of the aforementioned concept,
\begin{definition} \label{def:IIFOFP} \cite{zames1966input}
	System $H$ is considered to be instantaneously Incrementally Input Feed-Forward Output Feedback Passive (IIFOFP), if it is dissipative with respect to the input-output mapping $\omega(\cdot, \cdot)$,
	\begin{align*}
	&\omega(u_2-u_1,y_2-y_1)=(u_2-u_1)^T(y_2-y_1)\\&-\delta (y_2-y_1)^T(y_2-y_1)-\nu (u_2-u_1)^T(u_2-u_1),
	\end{align*}
	for some value $\nu \in R$ and $\delta \in R$ where $\nu\times\delta\leq0.25$.
\end{definition} 
\begin{remark}
	Although $\nu$ and $\delta$ may take negative or positive values, our goal is to design layers that have positive $\delta$'s. Further, a positive set of $\delta$ and $\nu$ for a layer implies IIFG stability and robustness for that layer. Nevertheless, as we will show, for the entire network to be robust and stable, the $\nu$ and $\delta$ for an individual layer can take any value as long as certain conditions are met.  $\nu$ and $\delta$ define a range of possible stability and robustness properties for a system. The lack of stability and robustness in one layer may be compensated for by an excess of robustness and stability in another layer \cite{khalil2002nonlinear}. In Subsection \ref{subsec:main3}, we provide an interpretation of the above relations and outline the implications behind the selection of $\nu$ and $\delta$.
\end{remark}
\begin{theorem} \cite{khalil2002nonlinear} \label{thm:L2Gam}
	If the nonlinear system $H$ is IIFOFP with $\delta>0$, then it is IIFG stable and robust with the finite gain $\gamma = \frac{1}{\delta}$.
\end{theorem}
Our goal in this paper is to connect Definition \ref{def:IIFOFP} to Definition \ref{def:FG} to achieve robustness and the property given in Theorem \ref{thm:L2Gam}. By enforcing the looser condition given in Definition \ref{def:IIFOFP}, where $\nu$ and $\delta$ can take a range of values locally at the layer level, we encourage a robust global behavior for the entire DNN as given in Definition \ref{def:FG}. According to Lyapunov theory for a layer to have the instantaneously IIFOFP property in Definition \ref{def:IIFOFP}, the following condition should hold for the input-output mapping of all the inputs $u_1$, $u_2$ fed to the layer and their respective output signals $y_1$, $y_2$: $\omega(u_2-u_1,y_2-y_1)>0$ \cite{khalil2002nonlinear}. If this holds, then the layer is dissipative with respect to the nonlinear relation given in Definition \ref{def:IIFOFP}. Our results will show how we can reach Definition \ref{def:FG} and robustness globally by enforcing Definition \ref{def:IIFOFP} locally. Lastly, we will use the following matrix properties in our proofs,
\begin{theorem} \label{thm:qd} \cite{kaszkurewicz2012matrix}
	A square matrix $A$ is a quasi-dominant matrix (diagonally dominant), if there exists a positive diagonal matrix $P=diag\{p_1,p_2,...,p_n\}$ such that $a_{ii}p_i\geq\sum_{j\neq i}|a_{ij}|p_j,~\forall i$, and/or $a_{jj}p_j\geq\sum_{i\neq j}|a_{ji}|p_i,~\forall j$. If these inequalities are met strictly, then the matrix is said to be strictly row-sum (or column-sum) quasi-dominant. If $P$ can be chosen as the identity matrix, then the matrix is said to be row- or column- diagonally dominant.
\end{theorem}
\begin{corollary} \label{cor:qdp} \cite{taussky1949recurring}
	Every symmetric quasi-dominant matrix is positive definite. 
\end{corollary}
\section{Theoretical Analysis and Main Results} \label{sec:main}
In our analysis, we treat each layer of the DNN as a nonlinear system as defined in Section \ref{sec:pre} i.e., $H_i$ for all layers $i=1,...,n$ (Fig. \ref{fig:dynsys_cascade}). A nonlinear system is defined as a layer in the network that accepts an input vector from the previous layer and produces an output vector with the size of the current layer after the weights, biases and activation functions are applied to the input signal.  We prove the conditions under which each layer $H_i$ is instantaneously IIFOFP with specific hyper-parameters $\delta_i$ and $\nu_i$. One can interpret the values of $\delta_i$ and $\nu_i$ as measures of robustness for the specific layer $i$. Our results place specific constraints on the weight matrices at the given layer based on the values of the hyper-parameters $\delta_i$ and $\nu_i$. We train the model and enforce these conditions during back-propagation. Consequently, we can show that the entire DNN is instantaneously IIFOFP and IIFG stable and robust. This means that the DNN maintains a level of robustness against adversarial changes added to the input for up to a specific adversarial $\ell_2$ norm $\epsilon$. This is because the changes in the output of the DNN are now bounded by the changes introduced to the input by the adversary. Robustness in this sense means that the adversary now needs to add a larger amount of noise to the input of the DNN in order to cause larger changes in the output of the DNN and affect the decision making process. Our approach improves robustness against adversarial changes added to the input signal and the output of a hidden layer before it is fed into the next layer. 

In our approach, we consider Leaky ReLU activation functions.  Our results and the Lyapunov theory suggest that Leaky ReLU is a more robust activation function than ReLu. We expand on this in Subsection \ref{subsec:main3}. $\Delta$ is a measure for intervention. $\Delta$ models the extent of adversarial noise introduced to the input by the adversary. The effects of the majority of $\epsilon$-based attacks of different norms such as the fast gradient method (FGM) and projected gradient descent (PGD) method may be modeled by $\Delta$ \cite{kurakin2016adversarial,madry2017towards,miyato2015distributional}. Given a DNN, we are interested in the (local) robustness of an arbitrary natural example $u$ by ensuring that all of its neighborhood has the same inference outcome. The neighborhood of $u$ may be characterized by an $\ell_2$ ball centered at $u$. We can define an adversarial input as follows,
\begin{definition} 
	Consider the input $u$ of the layer of size $n$, i.e., $u\in R^n$ and the perturbed input signal  $u+\Delta$ where $\Delta \in R^n$ is the attack vector that can take any value. The perturbed input vector $u+\Delta$ is within a $\Delta^0$-bounded $\ell_2$-ball centered at $u$ if $u+\Delta \in B_2(u,\Delta^0)$, where $B_2(u,\Delta^0):=\{u+\Delta|~||u+\Delta-u||_2=||\Delta||_2\leq\Delta^0\}$.
\end{definition}
Geometrically speaking, the minimum distance of a misclassified nearby instance to $u$ is the smallest adversarial strength needed to alter the DNN's prediction, which is also the largest possible robustness measure for $u$. We will use the conic behavior of the activation function, spectral norm of the weights and their relation to Lyapunov theory to train DNNs that are stable and robust against the adversary.
\subsection{The robustness analysis of each layer inside the deep neural network }  \label{subsec:main1}
Each layer of a DNN can be modeled as $ y_l = h_l(W_l u_{l} + b_l)$ for $l = 1,...,n$ for some $n>2$, where $u_{l} \in R^{n^\prime_{l-1}}$ is the input of the $l$-th layer, and $W_l \in R^{n^\prime_l \times n^\prime_{l-1}}$ and $b_l \in R^{n^\prime_l}$ are respectively the layer-wise weight matrix and bias vector applied to the flow of information from the layer $l-1$ to the layer $l$. $h_l : R^{n^\prime_{l-1}} \rightarrow R^{n^\prime_l}$ models the entire numerical transformation at the $l$-th layer including the (non-linear) activation functions. $n^\prime_{l-1}$ and $n^\prime_{l}$ represent the number of neurons in layers $l-1$ and $l$. For a set of weight and bias parameters, $\{W_l, b_l\}^{n}_{l=1}$, we can model the behavior of the entire DNN as $H_{\{W_l, b_l\}^{n}_{l=1}}(u_{1}) = y_{n}$ where $H_{\{W_l, b_l\}^{n}_{l=1}} : R^{n^\prime_1} \rightarrow R^{n^\prime_n}$ and $u_1$ is the initial input to the DNN. Given the training data set of size $K$, $(u_i , y_i )^{K}_{i=1}$, where $u_i \in R^{n^\prime_1}$ and $y_i \in R^{n^\prime_n}$, the loss function is defined as $\frac{1}{K} L(H_{\{W_l, b_l\}^{n}_{l=1}}(u_i), y_i)$, where $L$ is usually selected to be cross-entropy or the squared $\ell_2$-distance for classification and regression tasks, respectively. The model parameters to be learned is $x$.  We consider the problem of obtaining a model that is insensitive to the perturbation of the input. The goal is to obtain parameters, $\{W_l, b_l\}^{n}_{l=1}$, such that the $\ell_2$-norm of $h(u + \Delta) - h(u)$ is small, where $u \in R^{n^\prime_1}$ is an arbitrary vector and $\Delta \in R^{n^\prime_1}$ is an engineered perturbation vector with a small $\ell_2$-norm added by the adversary. 
To be more general and further investigate the properties of the layers, we assume that each activation function, modeled by $h_l$ is a modified version of element-wise ReLU called the Leaky ReLU: $h_l(y_l) = \max(y_l, a y_l)$, 
where $0<a<1$ (our results stand for simple ReLu as well). It follows that, to bound the variations in the output of the DNN by the variations in the input, it suffices to bound these variations for each $l \in \{1, ...,n\}$. Here, we consider that the attack variations $\Delta$ are added by the adversary into the initial input or the input of the hidden layers. This motivates us to consider a new form of regularization scheme, which is based on an individual layer's Lyapunov property. 
The first question we seek to answer is the following: what are the conditions under which a layer $l$ is IIFOFP with a positive $\delta_l$ and a $\nu_l$ that may take any value? In practice, it is best to train layers so that both $\nu_l$ and $\delta_l$ are positive, as this means a tighter bound and a more robust layer (Subsection \ref{subsec:main3}), however, this is not a necessary condition for our results.
\begin{theorem} \label{thm:bound}
	The numerical transformation at the hidden layer $l$ of the DNN as defined in Subsection \ref{subsec:main1} is instantaneously IIFOFP and consequently IIFG stable and robust, if the spectral norm of the weight matrix for the layer  satisfies the following condition,
	\begin{align*}
	&\rho(W_l) \leq \frac{1}{\delta_l^2}+\frac{2|\nu_l|}{\delta_l}
	\end{align*} 
	where  $\rho(W_l)$ is the spectral norm of the weight matrix at the layer $l$, and the hyper-parameters $\delta_l>0$ and $\nu_l$ meet the condition $\delta_l \times \nu_l \leq 0.25$.
\end{theorem}
\textit{Proof in Appendix \ref{app_1}} 
\begin{remark}
	It is important to note that the above theorem shows a relationship between the spectral norm of the weight matrix at the layer $l$ and instantaneously IIFG stability and robustness of the layer as defined in the Definition \ref{def:FG} and Theorem \ref{thm:L2Gam} through the hyper-parameters $\delta_l$ and $\nu_l$. Namely, a larger value for $\nu_l$ leads to a larger upper-bound for the spectral norm of the weight matrix at layer $l$. Larger values of $\delta_l$ however, have a reciprocal relation to the spectral norm of the weight matrix. The relationship between parameters $\delta_l$, $\nu_l$ and the spectral norm of the weight matrix may be utilized during the robust training of DNNs through the spectral regularization enforced at each layer.
\end{remark}
We can implement the above condition for each layer during the training of the network. If the above condition is met for each layer, then we can posit that the DNN is stable and robust in Lyapunov sense. The exact measure of stability and robustness depends on the selection of $\delta_l$ and $\nu_l$. The global effects of this choice are outlined in Subsection \ref{subsec:main2}. The extension of Theorem \ref{thm:bound} to convolutional layers follows in a similar pattern and is given in Appendix \ref{app_2}. Appendix \ref{app_3} includes the robustness analysis of ResNet building blocks. It is important to note that $\nu_l$ and $\delta_l$ are design hyper-parameters that are selected before the training starts. The only real conditions placed on the hyper-parameters are that $\delta_l$ should be positive and $\delta_l \times \nu_l \leq 0.25$. The exact implications of choosing the hyper-parameters and their effects on the robustness of the layer against adversarial noise are detailed in Subsection \ref{subsec:main3}. 
\subsection{The robustness analysis of the entire Deep Neural Network} \label{subsec:main2}
\begin{theorem} \label{thm:stab}
	Consider the cascade interconnection of hidden layers inside the DNN as given in Fig. \ref{fig:dynsys_cascade} where $n>2$, and each layer $H_l$ for $l=1,...,n$ is instantaneously IIFOFP with their respective $\nu_l$ and $\delta_l$ as defined in Theorem \ref{thm:bound}, i.e., for any two incremental inputs $u_{l1}$, $u_{l2}$ for the layer $l$ we have,
	\begin{align*}
	&\omega(u_{l2}-u_{l1},y_{l2}-y_{l1})=(u_{l2}-u_{l1})^T(y_{l2}-y_{l1})\\&-\delta_l (y_{l2}-y_{l1})^T(y_{l2}-y_{l1})-\nu_l (u_{l2}-u_{l1})^T(u_{l2}-u_{l1}),
	\end{align*}
	as the nonlinear input-output mapping of the layer where $\omega(u_{l2}-u_{l1},y_{l2}-y_{l1})>0$. Then the entire DNN is also instantaneously IIFOFP with the hyper-parameters $\nu$, $\delta$ and input-output mapping, 
	\begin{align*}
	&\omega(u_{2}-u_{1},y_{2}-y_{1})=(u_{2}-u_{1})^T(y_{2}-y_{1})\\&-\delta (y_{2}-y_{1})^T(y_{2}-y_{1})-\nu (u_{2}-u_{1})^T(u_{2}-u_{1}),
	\end{align*}
	where $u_1$ and $u_2$ are the initial inputs to the DNN and $y_1$ and $y_2$ are their respective output signals, if the matrix $-A$ is quasi-dominant, where $A$ is defined as,
	\begin{equation*} A = 
	\begin{bmatrix}
	\nu-\nu_1         & \frac{1}{2}          & 0              & \dots                       &  -\frac{1}{2}\\
	\frac{1}{2}        & -\delta_1 -\nu_2 & \frac{1}{2} & \dots                       & 0 \\
	\vdots              & \ddots               & \ddots      & \ddots                     &\vdots\\
	0                      &  \dots                & \frac{1}{2}& -\delta_{n-1} -\nu_n & \frac{1}{2}\\
	-\frac{1}{2}      & 0                       &  \dots       &  \frac{1}{2}              &\delta-\delta_n
	\end{bmatrix}.
	\end{equation*}
\end{theorem}
\textit{Proof in Appendix \ref{app_4}} 
\begin{remark} \label{remark}
	For the DNN to be IIFOFP with $\delta>0$ and $\nu>0$ and consequently stable and robust, we need to set the hyper-parameters for training such that $\delta_l>0$ for $l=1,...,n$, $\delta_n>\delta>0$, $\nu_1>0$, and $\nu_1>\nu>0$. The rest of $\nu_l$'s are selected such that the matrix $-A$ is quasi-dominant. Note that theoretically, the $\delta_l$'s or $\nu_l$'s for some hidden layers may take negative values, as long as the matrix $-A$ stays quasi-dominant, i.e., $\delta_l+\nu_{l+1}>1$ for $l=1,...,n-1$ and $\delta_{n-1}+\nu_{n}>1$. By selecting the hyper-parameters according to Theorem \ref{thm:stab}, one is indirectly setting the spectral regularization rule for each layer. Appendix \ref{app_5} details an example on the selection of these hyper-parameters.
\end{remark}
Theorem \ref{thm:stab} points to an interesting fact that the first and last hidden layer may have the largest effect on the robustness of the DNN. To keep the matrix $-A$ quasi-dominant, $\delta$ and $\nu$ have a direct dependence on the values of $\delta_n$ and $\nu_1$. Next, we can characterize a relationship between the incremental changes in the input signals of a DNN, i.e., $\Delta_1$, and their effects on the output of the DNN, i.e., $\Delta_n$. 
\begin{corollary} \label{corr}
	Consider the cascade interconnection of hidden layers inside the DNN as given in Fig. \ref{fig:dynsys_cascade} where $n>2$, if each layer $H_l$ is instantaneously IIFOFP  with their respective $\nu_i$  and $\delta_i$, and the DNN is trained to meet the conditions given in Theorem \ref{thm:stab}, then the entire DNN is also instantaneously IIFOFP with its respective $\nu$ and $\delta$ and the input-output mapping $\omega(u_{2}-u_{1},y_{2}-y_{1}) = (u_{2}-u_{1})^T(y_{2}-y_{1})-\delta (y_{2}-y_{1})^T(y_{2}-y_{1})-\nu (u_{2}-u_{1})^T(u_{2}-u_{1})$ where $\delta>0$. One can show that the variations in the final output of the entire DNN ($\Delta_n$) are upper-bounded (limited) by the variations in the input signal ($\Delta_1$) through the following relation,
	\begin{align*}
	&||\Delta_n||_2^2 \leq \left(\frac{1}{\delta^2}+\frac{2\nu}{\delta}\right)||\Delta_1||^2_2  = \left(\frac{1}{\delta^2}+\frac{2\nu}{\delta}\right) \epsilon^2
	\end{align*} 
	where  the design parameter $\delta$ and $\nu$ are both positive.
\end{corollary}
\textit{Proof in Appendix \ref{app_6}} 
\subsection{The conic interpretation of the proposed approach} \label{subsec:main3}

\setlength{\columnsep}{7.0pt}%
\setlength{\intextsep}{7.0pt}%
\begin{wrapfigure}{r}{0.25\textwidth}
	\centering
	\includegraphics[height=2.9cm]{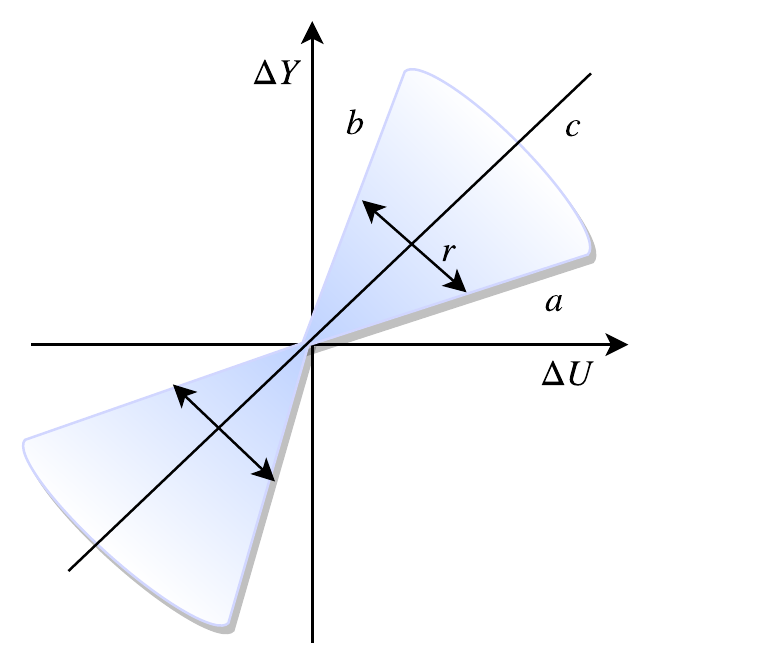}    
	\caption{A depiction of the interior conic behavior of a nonlinear system.}  
	\label{fig:conic}
\end{wrapfigure}
\textcite{zames1966input} was the first work that connected Lyapunov notion of stability and robustness to the conicity (bounded-ness) behavior of the input-output mapping of a nonlinear system. According to \cite{zames1966input}, a nonlinear numerical transformation is stable, if it produces bounded outputs for bounded inputs. The same nonlinear transformation is also robust, if it produces outputs that are insensitive to small changes added to the input. A stable and robust nonlinear numerical transformation then exhibits a conic behavior.

According to \cite{zames1966input}, a nonlinear transformation exhibits a conic behavior, if the mapping between the changes in the input and the respective changes in the output, given the nonlinear transformation, always fits inside a conic sector on the input-output plane i.e., a conic nonlinear numerical transformation in the Hilbert space is one whose input changes $\Delta u$ and output changes $\Delta y$ are restricted to some conic sector of the $\Delta U \times \Delta Y$ inner product space as given in Fig. \ref{fig:conic}. This conic behavior is usually defined by the center line of the cone $c$ and the radius $r$:
\begin{definition} \cite{zames1966input} \label{def:cone}
	A relation $H$ is interior conic, if there are real constants $r\geq0$ and c for which 
	$||\Delta y-c \Delta u||_{\ell_2} \leq r||\Delta u||_{\ell_2}$ is satisfied. 
\end{definition}
This is the exact behavior that we are encouraging for each layer of the DNN so that the outputs of each layer become insensitive to small changes in the input. In particular, we have $\Delta u=\Delta_1$, $\Delta y=\Delta_n$, $c=\frac{a+b}{2}$ and $r=\frac{b-a}{2}$ where $a$ and $b$ are the slopes of the lower and upper bounds of the cone, and $c$ and $r$ are the center and radius of the cone. One can show that,
\begin{align*}
&(\Delta_n-c\Delta_1)^T (\Delta_n-c\Delta_1) \leq r^2\Delta_1^T \Delta_1\\& \rightarrow
(\Delta_n-(\frac{a+b}{2})\Delta_1)^T (\Delta_n-(\frac{a+b}{2})\Delta_1)\\& \leq (\frac{b-a}{2})^2 \Delta_1^T \Delta_1  \rightarrow 
0\leq \Delta_1^T \Delta_n \\&- (\frac{1}{a+b})\Delta_n^T \Delta_n - (\frac{ba}{a+b})\Delta_1^T \Delta_1.
\end{align*}
Hence by selecting $\delta=\frac{1}{a+b}$ and $\nu=\frac{ba}{b+a}$, we are bounding the output changes by the changes in the input as depicted in Fig. \ref{fig:conic} and by that, we make the numerical transformations occurring at each layer of the DNN insensitive to small changes in the input. Particularly, a positive $\delta$ implies $b>0$ with a larger $\delta$ implying a smaller positive $b$ and a larger distance between the slope of the upper-bound of the cone and the $\Delta Y$ axis. This implies that $\Delta Y$ increases in a slower rate with increases in $\Delta U$. A positive $\nu$ implies $a>0$ with a larger $\nu$ implying a larger  $a$ and a larger distance between the lower-bound of the cone and the $\Delta U$ axis \cite{khalil2002nonlinear}. We are encouraging the pair $(\Delta_1, \Delta_n)$ to be instantaneously confined to a sector of the plane as depicted. The conic interpretation described here combined with the results given in the previous sections support the findings presented in \cite{qian2018l2,wong2018provable} that the use of Leaky Relu activation in the architecture of DNNs may contribute to robustness.
\section{Experiments} \label{sec:exp}
We validate our results by performing a diverse set of experiments on a variety of architectures (fully-connected, AlexNet, ResNet) and data sets (MNIST, CIFAR10, SVHN and ImageNet). Appendix \ref{app_15} contains the details on hyper-parameters and training process for the above architecture and data set combinations. The experiments are implemented in TensorFlow \cite{abadi2016tensorflow} and the code will be made readily available. We test the DNNs against the fast gradient method (FGM) attack \cite{goodfellow2014explaining} with Frobenius $\ell_2$ norm of $\epsilon \in [0.1, 0.4]$, and the iterative projected gradient descent (PGD) attack \cite{madry2017towards} with $100$ iterations, $\alpha=0.02 \epsilon$ and the same range of epsilons. Further, we show in Appendix \ref{app_11} that our approach provides improved robustness against the Carlini \& Wagner (C\&W) attack \cite{carlini2017towards}.

\begin{table}[!h]
\caption{The incremental output variations ($\Delta_n$) for the 3 layer forward-net given an attack strength and the bounds calculated according to Corollary \ref{corr} (MNIST, PGD attack)} \label{tbl:mnist_bound_results} \label{table}
\begin{adjustbox}{max size={\columnwidth}{\textheight}}
\begin{tabular}{@{}llll@{}}
\toprule
\multicolumn{1}{c}{}                           & \multicolumn{3}{c}{Mean output change $\|\Delta_n\|_2 \leq$ Lyapunov Bound}                                     \\ \cmidrule(l){2-4} 
\multicolumn{1}{c}{Lyapunov  Parameters} & \multicolumn{1}{c}{$\epsilon=0.1$} & \multicolumn{1}{c}{$\epsilon=0.2$} & \multicolumn{1}{c}{$\epsilon=0.3$} \\ \midrule
$\delta=0.89, \nu=0.28$                        & $0.244 \leq 0.435$                      & $0.490 \leq 0.615$                      & $0.736 \leq 0.753$                      \\
$\delta=0.95, \nu=0.26$                        & $0.202 \leq 0.407$                      & $0.406 \leq 0.576$                      & $0.610 \leq 0.706$                      \\
$\delta=1.1, \nu=0.22$                         & $0.170 \leq 0.352$                      & $0.340 \leq 0.497$                      & $0.511 \leq 0.609$                      \\
Base Model                                     & 1.233                              & 2.446                              & 3.641                              \\ \bottomrule
\end{tabular}
\end{adjustbox}
\end{table}

Our robust Lyapunov training method regularizes the spectral norm of a layer $l$ so that, $\rho(W_l)\leq\beta$ where $\beta= \frac{1}{\delta_l^2}+\frac{2|\nu_l|}{\delta_l}$. Given $n$ layers, one can pick $n$ different combinations of $(\rho_l, \nu_l)$ for a given data set and architecture as long as the conditions given in Theorem \ref{thm:bound}, Theorem \ref{thm:stab} and Remark \ref{remark} are met. Our proofs tell us that if we wish to constrain the adversarial perturbations, we should set the values of $(\rho_1, \nu_1)$ and $(\rho_n, \nu_n)$ with more care. The values of $(\rho_i, \nu_i), \text{ for } 1 < i < n$,  however, may be selected more freely to allow for a greater flexibility to learn while not giving the adversary further advantage. Appendix \ref{app_5} outlines the process for selecting the Lyapunov hyper-parameters according to our proofs. The different sets of Lyapunov design parameters used in our experiments are detailed in Appendix \ref{sec:experiment_design_details}. We represent a robust DNN with its global Lyapunov parameters $(\delta, \nu)$. It is important to note that the greater flexibility (higher expressiveness \cite{couellan2019coupling}) allowed for the intermediate layers leads to a better generalization on the clean and adversarial data sets. This is not true for DNNs trained by weight-decay or spectral norm regularization against a single threshold $\beta$. All the previous works on this subject \cite{farnia2018generalizable,qian2018l2,yoshida2017spectral,cisse2017parseval}, keep $\beta$ constant across layers. These harder constraints over-regularize and thus impair the DNN’s ability against attacks. Our results outlined in Appendix \ref{app_14} show that our Lyapunov DNNs are more robust and perform better in comparison to the aforementioned works.

Table \ref{table} details the effectiveness of our approach in bounding the incremental changes in the output of the DNN caused by the attack. The results show that the proposed bounds offered in Corollary \ref{cor:qdp} are met. Our bounds are never violated, however as $\epsilon$ gets larger, the output changes get closer to our proposed bounds. Given the empirically tightness of the bounds, one may be able to a-priori determine the worst case vulnerabilities against an attack for a network. We leave this to our future work in this area. When we consider the accuracy under attack, our Lyapunov approach dominates prior baselines. \textcite{farnia2018generalizable} for instance, obtained an accuracy of $62\%$ at $\epsilon=0.1$ with adversarial training on CIFAR10 under the $\ell_2$ PGD adversary with a lower number of iterations. Our approach obtains an accuracy of $\geq 73\%$ at $\epsilon=0.1$ and still dominates with an accuracy of $\geq 63\%$ at $\epsilon=0.4$ (Table \ref{tab:my-table6} in Appendix \ref{sec:experiment_results}). Fig. \ref{fig:cifar10_pgdm__fgsm_results} reports the CIFAR10 test accuracy under the iterative PGD and FGM attacks for different values of $\epsilon$. As noted in Section \ref{sec:main}, DNNs trained with larger global $(\delta, \nu)$ maintain their robustness in a more consistent way for larger $\epsilon$'s. This is because enforcing a larger global $(\delta, \nu)$ leads to DNNs with a more restricted conic behavior able to more effectively bound the negative effects of the adversarial noise. $\ell_2$ weight decay training seems ineffective against the attacks and cannot utilize adversarial training to improve its performance (Fig. \ref{fig:cifar10_pgdm_fgsm_results_adv} in Appendix \ref{sec:experiment_results}). Lastly, to show that our approach scales to larger architectures and data sets, Table \ref{tab:imagenet} represents our results for Lyapunov-based robust ResNet50 architectures trained on the ImageNet data set. A comprehensive set of results for a larger set of experiments is given in Appendix \ref{sec:experiment_results}. This appendix also includes our mathematical proofs on how Lyapunov based spectral regularization of the weights can improve the robustness of residual blocks.
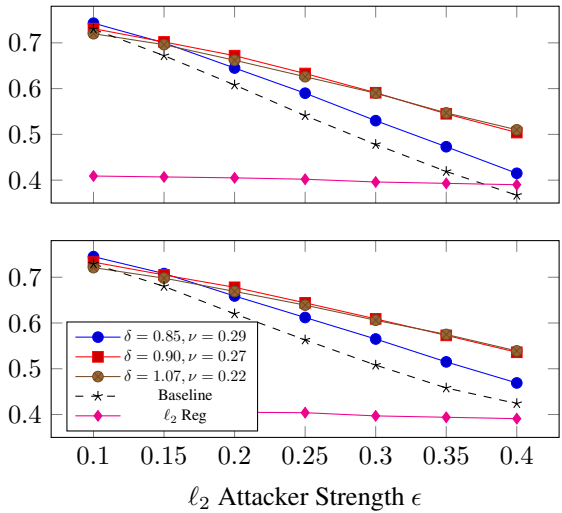
\begin{figure}[!htb]
	\begin{center}
		\centering
		\begin{tikzpicture}[]

		\begin{groupplot}[
		group style={
			group name=myplot, group size=1 by 2,
			vertical sep=0.5cm,
		},
		enlarge x limits=true,
		]
		\centering
		
		\nextgroupplot[
		height=4.2cm,
		width=1.0\columnwidth,
		xticklabels={,,},
		legend style={nodes={scale=0.6, transform shape}, at={(0.05,0.45)},anchor=west},
		legend pos=south west,
		legend columns=1,
		ymin=0.35,
		ymax=0.78,
		]

		\addplot+[] table [x=epsilon, y=d_086_v_29, ] {pgdm_cifar10_alexnet.dat};
		\addplot+[] table [x=epsilon, y=d_090_v_27, ] {pgdm_cifar10_alexnet.dat};
		\addplot+[] table [x=epsilon,y= d_107_v_22, ] {pgdm_cifar10_alexnet.dat};
		\addplot+[dashed] table [x=epsilon, y=no_reg, ] {pgdm_cifar10_alexnet.dat};
		\addplot+[magenta,mark options={magenta}] table [x=epsilon, y=l2_reg, green] {pgdm_cifar10_alexnet.dat};

		\nextgroupplot[
		height=4.2cm,
		width=1.0\columnwidth,
		xlabel={ $\ell_2$ Attacker Strength $\epsilon$},
		legend style={nodes={scale=0.6, transform shape}, at={(0.05,0.45)},anchor=west},
		legend pos=south west,
		legend columns=1,
		ymin=0.35,
		ymax=0.78,
		]

		\addplot+[] table [x=epsilon, y=d_086_v_29, ] {fgsm_cifar10_alexnet.dat};
		\addplot+[] table [x=epsilon, y=d_090_v_27, ] {fgsm_cifar10_alexnet.dat};
		\addplot+[] table [x=epsilon, y=d_107_v_22, ] {fgsm_cifar10_alexnet.dat};
		\addplot+[dashed] table [x=epsilon, y=no_reg, ] {fgsm_cifar10_alexnet.dat};
		\addplot+[magenta,mark options={magenta}] table [x=epsilon, y=l2_reg] {fgsm_cifar10_alexnet.dat};
		
		\legend{$\delta=0.85,\nu=0.29$ \\ 
			$\delta=0.90,\nu=0.27$\\
			$\delta=1.07,\nu=0.22$\\
			Baseline \\ 
			$\ell_2$ Reg\\
		}	
		
		\end{groupplot}
		
          [xshift=-1.0cm] 
          {\rotatebox{90}{Accuracy Under Attack}};
		
		\end{tikzpicture}
	\end{center}
	\caption{Accuracy of the DNN under PGD attack ($k=100$ iterations) (top) and FGM attack (bottom) using AlexNet on CIFAR10. Plots share the same legend and axis.
	} \label{fig:cifar10_pgdm__fgsm_results}
\end{figure}

\begin{table}[!htb]
		\centering
		\caption{Experiment results for ResNet50 trained on the ImageNet dataset under the FGM attack}
			\scalebox{0.65}{
	\begin{tabular}{|l|l|l|l|l|l|}
		\hline
		\multirow{2}{*}{Network Type}                                                                                                                                         & \multicolumn{4}{l|}{\begin{tabular}[c]{@{}l@{}}Accuracy on the Adversarial Test Dataset\\           (Attack Strength $\epsilon$)\end{tabular}} \\ \cline{2-5} 
		&                                                                                                            $\epsilon=0.1$                                   & $\epsilon=0.2$                               & $\epsilon=0.3$    & $\epsilon=0.4$                           \\ \hline
		$\delta=0.95$, $\nu=0.26$                                                                                                      &                                          \textbf{0.61}                              & \textbf{0.51}                     &\textbf{0.44}                 & \textbf{0.38}                                    \\ \hline
		$\delta=0.86$, $\nu=0.29$                                                                                                      &                                                  0.60                     & \textbf{0.51}                                        & 0.43                                     & 0.37                \\ \hline
		$\delta=0.74$, $\nu=0.33$                                                                                                       &                         0.60                                                 & 0.50                                          & 0.42                                     & 0.36                                     \\ \hline
		\begin{tabular}[c]{@{}l@{}} Base Training \end{tabular}                                  &                       0.57                                                  & 0.45                                        & 0.40                                       & 0.34                                      \\ \hline
		
		\begin{tabular}[c]{@{}l@{}}Base Training \\ with Freb. $\ell_2$ reg., $\lambda=0.01$\end{tabular}                                               &                           0.58                                               & 0.46                                          & 0.39                                       & 0.32                                    \\ \hline
	\end{tabular}
}
	\label{tab:imagenet}
\end{table}
\section{Conclusion} \label{sec:con}
In this paper, we analyzed the robustness of forward, convolutional, and residual layers against adversarial attacks based on Lyapunov theory of stability and robustness. We proposed a new robust way of training which improves robustness and allows for independent selection of the regularization parameters per layer. Our work bounds the layers' response to the adversary and gives more insights into how different architectures, activation functions, and network designs behave against attacks. 
{\small
\bibliographystyle{ieee_fullname}
\bibliography{AutBib,Raff}
}
\newpage
\appendix
\onecolumn
\section{Brief Primer on Control Theory}\label{controltheory}
Stability and Robustness of nonlinear systems have been studied in the field of control theory for more than a century. A nonlinear system is said to be stable, if given an input signal to the system, it can produce outputs that are expected by the user. A nonlinear system is said to be stable and robust, if given an input signal to the system and in the presence of external noise and disturbance, it can produce outputs that are expected by the user. In our work, we focus on bounded-input-bounded-output (BIBO) stability of nonlinear systems. Definition 1 in Preliminaries provides the mathematical definition for incremental BIBO stability. The Definition shows that the difference between two incremental outputs of the system should be bounded by the difference between their respective incremental inputs times $\gamma$. $\gamma$ is a positive constant (gain) indicating the extent of bounded-ness property of the system. A small $\gamma$ points to a more stable and robust system behavior. Our goal is to design a stable system that produces bounded outputs for the bounded signals it receives. In other words, we want to design a stable system that produces outputs close to the desired behavior defined by the user for the inputs from the domain of possible inputs. For instance, a DNN trained on the Imagenet data set is stable, if it produces desired classification decisions for any input image similar to the images in Imagenet data set. In our work, we focus on both stability and robustness. A robust stable DNN should be able to produce outputs close to the desired behavior defined by the user for the inputs from the domain possible inputs in cases where some amount noise has been added to the inputs. In our case, this noise is added by the adversary to the inputs. 

A nonlinear system, given an input, state and output, is mathematically defined as follows,\begin{equation*} \label{eq:dynsys}
H:
\begin{cases}
\dot{x}= f(x,u) 
&\\y= h(x,u),
\end{cases}
\end{equation*}
where  $x \in X \subseteq 
R^n$, $u \in U \subseteq
R^m$, and $y \in Y
\subseteq R^{k}$ are respectively the state, input and output of the system, and $X$, $U$ and $Y$ are the local state, input and output sub-spaces around the current operating points. The nonlinear mappings $f$ and $h$ model the relationship among the input signal $u$, the internal states of the system $x$ and the output signal $y$. For a DNN trained on the ImageNet data set, $U$ represents the images in the train and test data sets, $Y$ represents the domain of possible class decisions, and $X$ represents the domain of possible weight and bias values that could be assigned to the DNN's weights and biases. Further, $x$, $u$ and $y$ represents the possible realizations of these domains. As an example, $x$ can represent the values assigned to the weights and biases of the DNN after training and convergence.

$\dot{x}= f(x,u)$ represents the dynamic behavior of the DNN during training, where $u$ represents the training inputs to the DNN, and $f$ models the updates to the states of the DNN during the $t$ training iterations. This is also called the transient behavior of the system and $\dot{x}$ is a derivative taken over training iterations modeling the fact that the weights and biases are changing during training. In control theory, the transient behavior of a system is defined as the system's behavior before it reaches the stable equilibrium (steady-state). Lyapunov theory provides the foundation for defining stability and robustness for nonlinear systems solely based on their input-ouput behavior. This means that we can use Lyapunov results to define stability and robustness criteria for the system using the desired steady-state behavior $y=h(x, u_l)=h_l(\{W, B\}, u)$. This in return will determine the properties that the states (weights and biases) of the DNN should exhibit after the training is over. 

Lyapunov theory treats the input-output relationship as an energy based concept and proves that if the input-output behavior of a nonlinear systems meets the relationship given in Definition 2 then the system is stable and robust. Lyapunov theory states that a stable and robust nonlinear system dissipate energy and as a results they have to be dissipative with respective to the relationship given in Definition 2.

In our work, we treat each layer inside a DNN as a nonlinear system, and use the relationship given in Definition 2, to characterize the conditions which the weights and biases of the layer should meet for that layer to be stable and robust. We make sure that these conditions are met during training so that once the training is over, the layer is stable and robust. Further, we define conditions under which the entire cascade of the systems (the entire DNN) is stable and robust and we make sure that these condition are met after the training as well. As a result, we can provide a set of design options for spectral regularization of the weights for each layer inside the DNN, and bound the response of each layer and eventually the response of the entire network against adversarial attacks and the adversarial noise added to the inputs. Given our results, we can train DNNs that outperform other state-of-the-art robust solutions in the current literature.

\section{Proof of Theorem \ref{thm:bound}}  \label{app_1}
Given Definition \ref{def:IIFOFP}, we can show that for the two incremental inputs to the layer $l$, i.e., $u_{l,1}=u_{l}$ and $u_{l,2}=u_{l}+\Delta_{l}$ we have,
\begin{align}  \label{eq:relation1}
&\omega(u_{l} + \Delta_{l} - u_{l}, h(W_l [u_{l} + \Delta_{l}] + b_l)-h(W_l u_{l} + b_l)) \nonumber\\&=\Delta_{l}^T\Lambda_l W_l\Delta_{l}-\Delta_{l}^T(\nu_l I_l)\Delta_{l}-\Delta_{l}^TW_l^T\Lambda^T_l (\delta_l I_l)\Lambda_l W_l\Delta_{l}
\end{align}

where,
\begin{equation*}\Lambda_l = 
\begin{bmatrix}
1         &    0      & \dots   & \dots & \dots&           \\
0        & \ddots  & \ddots&          &        &\vdots \\
\vdots& \ddots  &  1        & 0       &        & \vdots  \\
\vdots&             & 0         & a         & \ddots&          \\
\vdots&             &            & \ddots & \ddots   &0 \\
& \dots    &  \dots  &           & 0          & a \\
\end{bmatrix}.
\end{equation*}

and $I_l$ is a diagonal identity matrix of size layer $l$. Further for a layer to be instantaneously IIFOFP with some $\nu_l$ and a positive $\delta_l$, one needs to show the following,
\begin{align*} 
&0\leq [u_{l}+ \Delta_{l}-u_{l}]^T[h_l(u_{l}+\Delta_{l})-h_l(u_{l})]\nonumber\\& - \delta_l [h_l(u_{l}+ \Delta_{l})-h_l(u_{l})]^T[h_l(u_{l}+ \Delta_{l})-h_l(u_{l})]\nonumber\\& - \nu_l [u_{l}+ \Delta_{l}-u_{l}]^T[u_{l}+ \Delta_{l}-u_{l}]
\end{align*}

Given (\ref{eq:relation1}) and for $\delta_l>0$, the above can be represented as,
\begin{align} \label{eq:relation3}
&0\leq (\frac{1}{2\delta_l}+|\nu_l|)||\Delta_{l}||^2_2 - \frac{\delta_l}{2} ||\Lambda_l W_l\Delta_{l}||_2^2
\end{align} 

Further, the following properties hold,  $||\Lambda_l W_l\Delta_{l}||_2^2< ||\Lambda_l||_2^2 ||W_l||_2^2||\Delta_{l}||_2^2$, and since $0<a<1$, we have $||\Lambda_l||_2^2<\lambda_{max}(\Lambda_l^T\Lambda_l)\leq1$ where $\lambda_{max}(.)$ stands for the largest eigenvalue (singular value) of a matrix. Also we can show that $\rho(W_l)<||W_l||_2^2<\rho(W_l)+\sigma$ where $\rho(W_l)$ is the spectral radius of $W_l$ and $\sigma$ is a small positive number. Simplifying (\ref{eq:relation3}) further we have,
\begin{align} \label{eq:relation4}
&0\leq (\frac{1}{2\delta_l}+|\nu_l|)||\Delta_{l}||^2_2 - \frac{\delta_l}{2} ||\Lambda_l W_l\Delta_{l}||_2^2\nonumber\\&
\leq (\frac{1}{2\delta_l}+|\nu_l|)||\Delta_{l}||^2_2 - \frac{\delta_l}{2} ||\Lambda_l||_2^2 ||W_l||_2^2||\Delta_{l}||_2^2\nonumber\\&
\leq(\frac{1}{2\delta_l}+|\nu_l|)||\Delta_{l}||^2_2 - \frac{\delta_l}{2} [(\rho(W_l)+\sigma)||\Delta_{l}||_2^2]\nonumber\\&
\approx[\frac{1}{2\delta_l}+|\nu_l| - \frac{\delta_l}{2} \rho(W_l)]||\Delta_{l}||_2^2
\end{align} 

For the relation (\ref{eq:relation4}) to be positive, the term inside the bracket needs to be positive. This gives us a measure for spectral regularization of the weights between each two hidden layers of a DNN. For layer $l$ we have,
\begin{align*} 
&\rho(W_l) \leq \frac{1}{\delta_l^2}+\frac{2|\nu_l|}{\delta_l},
\end{align*} 

which proves the theorem.

\section{Robustness analysis of a convolutional layer}  \label{app_2}
The transformations before the activation functions at the convolutional layers are linear, and the same isomorphism as for the linear layers can be exploited for the convolutional layers to have the same final relationships as given in Theorem \ref{thm:bound}. More specifically, we can follow the steps given in \cite{gouk2018regularisation} to define the transformation occurring at a convolutional layer $l$ for output feature $i$ with any padding and stride design as, 
\begin{align*} 
&\phi^{conv}_{l,i}(u_{l,i}) =\sum_{j=1}^{M_{l-1}}f_{j,i}\ast u_{l,j,i} +b_{l,i}
\end{align*} 

Each $f_{j,i}$ is a filter applied to the input feature and each $u_{l,j,i}$ is an input feature map from the previous layer. $b_{l,i}$ is an appropriately shaped biased tensor adding the same value to every element resulting from the convolutions. $M_{l-1}$ is the number of feature maps in the previous layer. One can represent the above relation as a matrix-vector multiplication by defining the serialized version of the input, $U_{l,i} = [u_{l,1,i}, ..., u_{l,M_{l-1},i}]$ and then representing the filter coefficients in the form of a doubly block circulant matrix \cite{sedghi2018singular}. Specifically, if $F_{j,i}$ is a matrix that encompasses convolution of $f_{j,i}$ with the $j$-th feature map in a vector form, then to represent convolutions associated with different input feature maps and the same output feature map i.e., $f_{j,i}$'s over $M_{l-1}$ input features, one can horizontally concatenate the filter matrices to define $F_i = [F_{1,i}, F_{2,i}, ... ,F_{M_{l-1},i}]$. Then the complete transformation performed by a convolutional layer to generate $M_l$ output feature maps can be represented as,
\begin{align*} 
&h_l(U_{l}) = h_l(W U_{l} + B_l)
\end{align*} 

where,
\begin{align*} 
&W =  \begin{bmatrix}
F_{1,1}& \dots & F_{M_{l-1},1}\\
\vdots & \ddots&\vdots\\
F_{1,M_l}\ & \dots&  F_{M_{l-1},M_l}
\end{bmatrix}
\end{align*} 

and where vector $B_l$ is the larger version of $b^l_i$'s for all input feature maps and $U_{l}=[U_{l,1},...,U_{l,M_{l-1}}]$. Consequently, the spectral norm of $W$ should meet the conditions given in Theorem \ref{thm:bound} for the layer $l$ to be IIFOFP and IIFG stable with bounded incremental outputs. We use the power iteration method to estimate the spectral norm of the weight matrix at a specific layer during training as proposed in \cite{farnia2018generalizable}. Further, the pooling layers inside a DNN do not affect the conic behavior of the sub-systems given the properties of conic systems as described in \cite{zames1966input}. More specifically, depending on how the pooling layer is designed, an adjustment is made to the $\ell_2$ norm of the incremental output changes for the sub-system containing the pooling layer. Max (average) pooling decreases the norm of the changes and as a a result the conditions given in (\ref{eq:relation3}) is still met after the pooling layer. 

\section{Robustness analysis of a ResNet building block}  \label{app_3}
The following represents the  input-output mapping of a building block $l$ for incremental inputs $u_{l2}, u_{l1}$ and outputs $y_{l2}, y_{l1}$ in a ResNet layer \cite{he2016deep}, where $y_{li} = u_{li} + \mathcal{F}(u_{li}, \{W_l\})$,
\begin{align*}
&M_l=(u_{l2}-u_{l1})^T(y_{l2}-y_{l1})-\delta_l (y_{l2}-y_{l1})^T(y_{l2}-y_{l1})-\nu_l(u_{l2}-u_{l1})^T(u_{l2}-u_{l1})
\\&=(u_{l2}-u_{l1})^T(u_{l2} +\mathcal{F}_2-u_{l1} - \mathcal{F}_1)-\delta_l (u_{l2} + \mathcal{F}_2- u_{l1} - \mathcal{F}_1)^T(u_{l2} + \mathcal{F}_2- u_{l1} - \mathcal{F}_1)\\&-\nu_l(u_{l2}-u_{l1})^T(u_{l2}-u_{l1})
\\&=(u_{l2}-u_{l1})^T(\mathcal{F}_2- \mathcal{F}_1)-\delta_l (u_{l2} + \mathcal{F}_2- u_{l1} - \mathcal{F}_1)^T(u_{l2} + \mathcal{F}_2- u_{l1} - \mathcal{F}_1)\\&+(1-\nu_l)(u_{l2}-u_{l1})^T(u_{l2}-u_{l1})
\\&=(1-2\delta_l)(u_{l2}-u_{l1})^T(\mathcal{F}_2- \mathcal{F}_1)-\delta_l (\mathcal{F}_2 - \mathcal{F}_1)^T(\mathcal{F}_2- \mathcal{F}_1)\\&-(\nu_l+\delta_l-1)(u_{l2}-u_{l1})^T(u_{l2}-u_{l1})
\\&=(u_{l2}-u_{l1})^T(\mathcal{F}_2- \mathcal{F}_1)-\frac{\delta_l }{1-2\delta_l}(\mathcal{F}_2 - \mathcal{F}_1)^T(\mathcal{F}_2- \mathcal{F}_1)-\frac{\nu_l+\delta_l-1}{1-2\delta_l}(u_{l2}-u_{l1})^T(u_{l2}-u_{l1})
\end{align*} 

where we have replaced  $\mathcal{F}_i(u_{li}, \{W_l\})$ with $ \mathcal{F}_i$ for simplicity. The above relation tells us that the feed-forward connection from the input to the output degrades the robustness of the ResNet block. This is because for a ResNet block $l$ to be stable and robust, the following should hold for the Lyapunov parameters of the block excluding the feed-forward connection: $0<\delta_l <\frac{1}{2}$ and $\nu_l+\delta_l>1$ where $\delta_l\times\nu_l<0.25$. So that the entire block can have the robustness properties: $\delta_l^\prime=\frac{\nu_l+\delta_l-1}{1-2\delta_l}$ and $\nu_l^\prime=\frac{\nu_l+\delta_l-1}{1-2\delta_l}$. This means that to maintain robustness and stability for a ResNet block, one needs to enforce stricter conditions on the Lyapunov design hyper-parameters of the sub-layers inside the block and consequently the spectral norm of the weights during the training of the DNN. In a sense, this means that a ResNet block is less robust against adversarial attacks in comparison to a simple feedforward or convolutional layer. This is expected because the output of a ResNet block consists of the block's output and the input signal fed into the block. This means that if the input signal is perturbed by adversarial noise, then under this architecture, the adversarial noise can easily propagate to the output of the block and throughout the DNN. The negative effects of feed-forward connections on robustness of nonlinear systems have been explored in control theory \cite{khalil2002nonlinear}.

\section{Proof of Theorem \ref{thm:stab}}  \label{app_4}
The input-output mapping of each layer can be represented as $M_l=(u_{l2}-u_{l1})^T(y_{l2}-y_{l1})-\delta_l (y_{l2}-y_{l1})^T(y_{l2}-y_{l1})-\nu_l(u_{l2}-u_{l1})^T(u_{l2}-u_{l1})>0$ for layers $l=1,..,N$, one needs to show that,
\begin{align*} 
&0\leq \sum_{l=1}^{n} M_l \leq (u_{2}-u_{1})^T(y_{2}-y_{1})-\delta (y_{2}-y_{1})^T(y_{2}-y_{1})-\nu (u_{i2}-u_{1})^T(u_{2}-u_{1}).
\end{align*} 

The summation $\sum_{l=1}^{n} M_l$ is positive if the sub-layers are trained according to Theorem \ref{thm:bound} and as a result one can show that the above relation is equivalent to,
\begin{align} \label{eq:relation6}
&\sum_{l=1}^{n} M_l - (u_{2}-u_{1})^T(y_{2}-y_{1})+\delta (y_{2}-y_{1})^T(y_{2}-y_{1})+\nu (u_{2}-u_{1})^T(u_{2}-u_{1})\leq0.
\end{align} 

We can define the following matrices,
\begin{equation*} A_1 = 
\begin{bmatrix}
-\nu_1& 0 & \dots & 0 & -\frac{1}{2}\\
\frac{1}{2}& -\nu_2 & \ddots & & 0  \\
\vdots & \ddots&  \ddots  & \ddots & \vdots \\
0 & \dots &  \frac{1}{2} & -\nu_n & 0 \\
0 & \dots & 0 & \frac{1}{2}& \delta
\end{bmatrix},
\end{equation*}

and,
\begin{equation*} A_2 = 
\begin{bmatrix}
\nu& 0 & \dots & 0 & -\frac{1}{2}\\
\frac{1}{2}& -\delta_1 & \ddots & & 0  \\
0 &\frac{1}{2}& -\delta_2  & \ddots & \vdots \\
\vdots & \ddots & \ddots & \ddots & 0 \\
0 & \dots & 0 & \frac{1}{2}& -\delta_n
\end{bmatrix}.
\end{equation*}

As a result, (\ref{eq:relation6}) may be represented as,
\begin{equation*} 
[u^T y^T] (A_1^T+A_2)[u^T y^T]^T=[u^T y^T] A[u^T y^T]^T,
\end{equation*}

where,
\begin{equation*} A=A_1^T+A_2 = 
\begin{bmatrix}
\nu-\nu_1         & \frac{1}{2}          & 0              & \dots                       &  -\frac{1}{2}\\
\frac{1}{2}        & -\delta_1 -\nu_2 & \frac{1}{2} & \dots                       & 0 \\
\vdots              & \ddots               & \ddots      & \ddots                     &\vdots\\
0                      &  \dots                & \frac{1}{2}& -\delta_{n-1} -\nu_n & \frac{1}{2}\\
-\frac{1}{2}      & 0                       &  \dots       &  \frac{1}{2}              &\delta-\delta_n
\end{bmatrix}.
\end{equation*}

According to Corollary \ref{cor:qdp}, if $-A$ is quasi-dominant, then$-A$ is positive definite, which means that $A$ is negative definite and we have $[u^T y^T] A[u^T y^T]^T\leq 0$ and the relation given in (\ref{eq:relation6}) is met. As a result the DNN is instantaneously IIFOFP with indices $\delta$ and $\nu$. For this to hold, the only condition is for the hyper-parameters to be selected such that the matrix $-A$ is quasi-dominant. For the case that we are interested in, we need the hyper-parameters to be selected such that $\delta>0$, $\nu>0$ where $\delta_n>\delta>0$ and $\nu_1>\nu>0$, $\delta_i>0$ for $i=1,...,n$ and $\nu_i$ for $i=1,...,n$ are selected such that the matrix $-A$ is quasi-dominant. 

\section{An example on how the Lyapunov parameters are selected}  \label{app_5}
Here, we provide an example of parameter selection for the fully connected DNN used in some of our experiments with the global Lyapunov property of $\delta=1.0$, $\nu=0.24$. For a 3 layer fully connected forward-net, the following Lyapunov hyper-parameters should be selected: $\delta_1, \nu_1$ ($\rho(W_1)$), $\delta_2, \nu_2$ ($\rho(W_2)$), $\delta_3, \nu_3$ ($\rho(W_3)$), which then determine the global Lyapunov properties: $\delta, \nu$. The following conditions should be met for the matrix $-A$ to be quasi-dominant: $\nu<\nu_1$, $\delta<\delta_3$, $\delta_1+\nu_2>1$, $\delta_2+\nu_3>1$. The matrix $-A$ is:
\begin{equation*} -A = 
\begin{bmatrix}
\nu_1-\nu          & -\frac{1}{2}          & 0                       & \frac{1}{2}\\
-\frac{1}{2}        & \delta_1 + \nu_2 & -\frac{1}{2}                 & 0 \\
0      & -\frac{1}{2}          & \delta_2+ \nu_3                & -\frac{1}{2} \\
\frac{1}{2}         & 0                       &  -\frac{1}{2}              &\delta_3-\delta
\end{bmatrix}.
\end{equation*}

For $\delta=1.0$, we have a range of choices for $\delta_3$ ($\delta<\delta_3$), we select $\delta_3=1.08$ which is a robust choice due to its relatively large value (Subsection \ref{subsec:main3}).This gives us the condition $\nu_3<0.25/1.08=0.231$,  where $\nu_3=0.23$ is selected to meet the condition. As a result, the allowed range for spectral regularization for the last layer is $\rho(W_3)< \frac{1}{\delta_3^2}+\frac{2|\nu_3|}{\delta_3}=1.283$. In a similar manner for $\nu=0.24$, we have a range of choices for $\nu_1$ ($\nu<\nu_1$), we select $\nu_1=0.27$. This gives us the condition $\delta_1<0.25/0.27=0.925$, where $\delta_1=0.92$ is selected to meet the condition.  As a result, the allowed range for spectral regularization of the first layer becomes $\rho(W_1)< \frac{1}{\delta_1^2}+\frac{2|\nu_1|}{\delta_1}=1.768$. Lastly, the values of $\nu_2$ and $\delta_2$ should be selected such that $\nu_2>0.08$ and $\delta_2>0.769$, we select $\delta_2=0.78$ which gives us $\nu_2=0.32$ and the spectral regularization range for the second layer becomes, $\rho(W_2)< \frac{1}{\delta_2^2}+\frac{2|\nu_2|}{\delta_2}=2.464$. For our experiment, we use cross-validation to select the following spectral regularization set $[1.76, 2.46, 1.01]$. The matrix $-A$ becomes,
\begin{equation*} -A = 
\begin{bmatrix}
0.03         & -\frac{1}{2}          & 0                       & \frac{1}{2}\\
-\frac{1}{2}        & 1.24 & -\frac{1}{2}                 & 0 \\
0      & -\frac{1}{2}          & 1.11            & -\frac{1}{2} \\
\frac{1}{2}         & 0                       &  -\frac{1}{2}              &0.08
\end{bmatrix},
\end{equation*}

which is a quasi-dominant (diagonally dominant) matrix and positive definite with the spectral norm of $1.791$ and Frobenius norm of $2.185$.

\section{Proof of Corollary \ref{corr}}  \label{app_6}
Given Theorem \ref{thm:stab}, the definitions for the following vectors, $\Delta_n=y_2-y_1$ representing the changes in the output of the DNN, $\Delta_1 =u_1 + \Delta_1- u_1$ representing the changes in the input of the DNN injected by the attacker and according to Theorem \ref{thm:bound}, we have,
\begin{align*}
&0 \leq\Delta_1 ^T\Delta_n- \Delta_n^T\delta I\Delta_n-\Delta_1 ^T\nu I\Delta_1 
\end{align*}

Given that $\delta>0$ and  $\nu>0$, the above can be represented as,
\begin{align*}
&0\leq (\frac{1}{2\delta}+\nu)||\Delta_1 ||^2_2 - \frac{\delta}{2} ||\Delta_n||_2^2
\end{align*} 

Finally, if we move the appropriate terms to the left side of the above inequalities we have,
\begin{align*} 
&||\Delta_n||_2^2 \leq (\frac{1}{\delta^2}+\frac{2\nu}{\delta})||\Delta_1||^2_2.
\end{align*} 

\section{Experiment design and hyper-parameter details } \label{sec:experiment_design_details}
This appendix includes all hyper-parameters used in the experiments. Please note that, each layer has their own $\delta$ and $\nu$ associated with them. The pair ($\delta$, $\nu$) then determine the level of spectral regularization, given in parenthesis,  enforced at the layer during the training of network.  The $\delta$'s and $\nu$'s are selected according to the conditions presented in Definition \ref{def:IIFOFP}, Theorem \ref{thm:L2Gam}, Theorem \ref{thm:bound}, Theorem \ref{thm:stab} and Corollary \ref{cor:qdp}. The selections of the pairs ($\delta$, $\nu$) for the layers then lead to a global Lyapunov pair ($\delta$, $\nu$) for the entire network which then later is used to present a specific network in the tables.
\subsection{The Lyapunov design parameters for Forward-Net (trained on the MNIST dataset)}  \label{app_8}
\begin{table}[H]
		\centering
	\caption{The hyper-parameters for the fully connected Forward-Net (3 layers of sizes [50, 20, 10]) used in the experiments \newline}
	\scalebox{.87}{
		\begin{tabular}{|l|l|l|l|l|}
			\hline
			\begin{tabular}[c]{@{}l@{}}Forward-Net\end{tabular} & Layer 1 (linear)                                                            & Layer 2 (linear)                                                            & Layer 3 (linear)                                                                        & \begin{tabular}[c]{@{}l@{}}Global Lyapunov\\ Property\end{tabular}                                      \\ \hline
			\multirow{7}{*}{\begin{tabular}[c]{@{}l@{}} Design\\ Parameters \end{tabular}   }                              & \begin{tabular}[c]{@{}l@{}}$\delta$, $\nu$ \\ (spect. norm reg.)\end{tabular} & \begin{tabular}[c]{@{}l@{}}$\delta$, $\nu$ \\ (spect. norm reg.)\end{tabular} & \begin{tabular}[c]{@{}l@{}}$\delta$, $\nu$ \\ (spect. norm reg.)\end{tabular}             & $\delta$, $\nu$                                                                                           \\ \cline{2-5} 
			& \begin{tabular}[c]{@{}l@{}}$\delta=0.86$, $\nu=0.29$\\ (1.83)\end{tabular}    & \begin{tabular}[c]{@{}l@{}}$\delta=0.90$, $\nu=0.27$\\ (1.83)\end{tabular}    & \begin{tabular}[c]{@{}l@{}}$\delta=0.90$, $\nu=0.27$\\ (1.83)\end{tabular}                & \begin{tabular}[c]{@{}l@{}}$\delta=0.89$,  $\nu=0.28$\end{tabular}                                      \\ \cline{2-5} 
			& \begin{tabular}[c]{@{}l@{}}$\delta=0.92$, $\nu=0.27$\\ (1.76)\end{tabular}    & \begin{tabular}[c]{@{}l@{}}$\delta=0.98$, $\nu=0.25$\\ (1.55)\end{tabular}    & \begin{tabular}[c]{@{}l@{}}$\delta=0.96$, $\nu=0.26$\\ (1.62)\end{tabular}                & \begin{tabular}[c]{@{}l@{}}$\delta=0.95$,  $\nu=0.26$\end{tabular}                                      \\ \cline{2-5} 
			& \begin{tabular}[c]{@{}l@{}}$\delta=0.92$, $\nu=0.27$\\ (1.76)\end{tabular}     & \begin{tabular}[c]{@{}l@{}}$\delta=0.78$, $\nu=0.32$\\ (2.46)\end{tabular}     & \begin{tabular}[c]{@{}l@{}}$\delta=1.08$, $\nu=0.23$\\ (1.01)\end{tabular} & \begin{tabular}[c]{@{}l@{}}$\delta=1.0$, $\nu=0.24$\end{tabular}                                       \\ \cline{2-5} 
			& N/A                                                                         & N/A                                                                         & N/A                                                                                     & \begin{tabular}[c]{@{}l@{}}None\\ (Base Training with Freb. \\ $\ell_2$ reg., $\lambda=0.01$)\end{tabular} \\ \cline{2-5} 
			& N/A                                                                         & N/A                                                                         & N/A                                                                                     & \begin{tabular}[c]{@{}l@{}}None\\ (Base Training with Freb.\\ $\ell_2$ reg., $\lambda=0.05$)\end{tabular}     \\ \cline{2-5} 
			& N/A                                                                         & N/A                                                                         & N/A                                                                                     & \begin{tabular}[c]{@{}l@{}}None\\ (Base Training with Freb.\\ $\ell_2$ reg., $\lambda=0.1$)\end{tabular}      \\ \hline
		\end{tabular}
	}
	\label{tab:my-table2}
\end{table}

\subsection{The Lyapunov design parameters for AlexNet (trained on the CIFAR-10 dataset)}  \label{app_7}
\begin{table}[H]
		\centering
	\caption{The hyper-parameters for the AlexNet architecture used in the experiments \newline}
	\scalebox{0.64}{
		\begin{tabular}{|l|l|l|l|l|l|l|}
			\hline
			AlexNet:                           & Layer 1 (conv.)                                                             & Layer 2 (conv.)                                                             & Layer 3 (conv.)                                                             & Layer 4 (linear)                                                            & Layer 5 (linear)                                                            & \begin{tabular}[c]{@{}l@{}}Global Lyapunov\\ Property\end{tabular}                                           \\ \hline
			\multirow{7}{*}{\begin{tabular}[c]{@{}l@{}} Design\\ Parameters \end{tabular}   } & \begin{tabular}[c]{@{}l@{}}$\delta$, $\nu$ \\ (spect. norm reg.)\end{tabular} & \begin{tabular}[c]{@{}l@{}}$\delta$, $\nu$\\  (spect. norm reg.)\end{tabular} & \begin{tabular}[c]{@{}l@{}}$\delta$, $\nu$\\  (spect. norm reg.)\end{tabular} & \begin{tabular}[c]{@{}l@{}}$\delta$, $\nu$ \\ (spect. norm reg.)\end{tabular} & \begin{tabular}[c]{@{}l@{}}$\delta$, $\nu$ \\ (spect. norm reg.)\end{tabular} & $\delta$, $\nu$                                                                                                \\ \cline{2-7} 
			& \begin{tabular}[c]{@{}l@{}}$\delta=0.74$, $\nu=0.33$\\ (2.52)\end{tabular}      & \begin{tabular}[c]{@{}l@{}}$\delta=0.74$, $\nu=0.33$\\ (2.52)\end{tabular}      & \begin{tabular}[c]{@{}l@{}}$\delta=0.74$, $\nu=0.33$\\ (2.52)\end{tabular}      & \begin{tabular}[c]{@{}l@{}}$\delta=0.74$, $\nu=0.33$\\ (2.52)\end{tabular}      & \begin{tabular}[c]{@{}l@{}}$\delta=0.86$, $\nu=0.29$\\ (2.02)\end{tabular}      & \begin{tabular}[c]{@{}l@{}}$\delta=0.85$, $\nu=0.29$\end{tabular}                                             \\ \cline{2-7} 
			& \begin{tabular}[c]{@{}l@{}}$\delta=0.86$, $\nu=0.29$\\ (2.02)\end{tabular}      & \begin{tabular}[c]{@{}l@{}}$\delta=0.86$, $\nu=0.29$\\ (2.02)\end{tabular}      & \begin{tabular}[c]{@{}l@{}}$\delta=0.86$, $\nu=0.29$\\ (2.02)\end{tabular}      & \begin{tabular}[c]{@{}l@{}}$\delta=0.86$, $\nu=0.29$\\ (2.02)\end{tabular}      & \begin{tabular}[c]{@{}l@{}}$\delta=0.92$, $\nu=0.27$\\ (1.76)\end{tabular}      & \begin{tabular}[c]{@{}l@{}}$\delta=0.90$, $\nu=0.27$\end{tabular}                                             \\ \cline{2-7} 
			& \begin{tabular}[c]{@{}l@{}}$\delta=0.92$, $\nu=0.27$\\ (1.76)\end{tabular}      & \begin{tabular}[c]{@{}l@{}}$\delta=0.78$, $\nu=0.32$\\ (2.46)\end{tabular}      & \begin{tabular}[c]{@{}l@{}}$\delta=0.78$, $\nu=0.32$\\ (2.46)\end{tabular}      & \begin{tabular}[c]{@{}l@{}}$\delta=0.78$, $\nu=0.32$\\ (2.46)\end{tabular}      & \begin{tabular}[c]{@{}l@{}}$\delta=1.08$, $\nu=0.23$\\ (1.01)\end{tabular}      & \begin{tabular}[c]{@{}l@{}}$\delta=1.07$, $\nu=0.22$\end{tabular}                                             \\ \cline{2-7} 
			& N/A                                                                         & N/A                                                                         & N/A                                                                         & N/A                                                                         & N/A                                                                         & \begin{tabular}[c]{@{}l@{}}None\\ (Base training)\end{tabular}                                               \\ \cline{2-7} 
			& N/A                                                                         & N/A                                                                         & N/A                                                                         & N/A                                                                         & N/A                                                                         & \begin{tabular}[c]{@{}l@{}}None\\ (Base Training  with\\ Freb. $\ell_2$ reg., $\lambda=0.01$)\end{tabular}                          \\ \cline{2-7} 
			& \begin{tabular}[c]{@{}l@{}}N/A, N/A\\ (1.0)\end{tabular}                    & \begin{tabular}[c]{@{}l@{}}N/A, N/A\\ (1.0)\end{tabular}                    & \begin{tabular}[c]{@{}l@{}}N/A, N/A\\ (1.0)\end{tabular}                    & \begin{tabular}[c]{@{}l@{}}N/A, N/A\\ (1.0)\end{tabular}                    & \begin{tabular}[c]{@{}l@{}}N/A, N/A\\ (1.0)\end{tabular}                    & \begin{tabular}[c]{@{}l@{}}None\\ (Training with spectral\\ reg. \textless{}1 for each layer)\end{tabular} \\ \hline
		\end{tabular}
	}
	\label{tab:my-table1}
\end{table}

\subsection{The Lyapunov design parameters for the ResNet architecture (trained on the SVHN dataset)} \label{app_13}
\begin{sidewaystable} 
	\begin{table}[H]
			\centering
		\caption{The hyper-parameters for the ResNet architecture (conv. layer, 4 blocks of size 2, linear layer) used in the experiments 
		\newline}
		\scalebox{.55}{
			\begin{tabular}{|l|l|l|l|l|l|l|l|}
				\hline
				ResNet          & Layer 1 (conv.)           & Block 1          & Block 2                & Block 3          & \multicolumn{1}{c|}{Block 4}     & Layer 5 (linear)          & \begin{tabular}[c]{@{}l@{}}Global Lyapunov\\ Property\end{tabular}       \\ \hline
				\multirow{5}{*}{\begin{tabular}[c]{@{}l@{}} Design\\ Parameters \end{tabular}   } & $\delta$, $\nu$    & \begin{tabular}[c]{@{}l@{}}$\delta$, $\nu$\\ $\delta_1$, $\nu_1$, $\delta_2$, $\nu_2$  \\ (spect. norm reg.), (spect. norm reg.)\end{tabular}         & \begin{tabular}[c]{@{}l@{}} $\delta$, $\nu$, \\$\delta_1$, $\nu_1$, $\delta_2$, $\nu_2$\\ (spect. norm reg.), (spect. norm reg.) \end{tabular}  & \begin{tabular}[c]{@{}l@{}}  $\delta$, $\nu$,\\ $\delta_1$, $\nu_1$, $\delta_2$, $\nu_2$\\ (spect. norm reg.), (spect. norm reg.)  \end{tabular} & \begin{tabular}[c]{@{}l@{}}$\delta$, $\nu$\\ $\delta_1$, $\nu_1$, $\delta_2$, $\nu_2$ \\ (spect. norm reg.), (spect. norm reg.)\end{tabular}          & \begin{tabular}[c]{@{}l@{}}$\delta$, $\nu$\\ (spect. norm reg.)\end{tabular} & $\delta$, $\nu$                                                                                \\ \cline{2-8} 
				& \begin{tabular}[c]{@{}l@{}}$\delta=0.84$, $\nu=0.29$\\ (2.08)\end{tabular} & \begin{tabular}[c]{@{}l@{}}$\delta=1.16$, $\nu=0.20$\\ $\delta_1=0.351$, $\nu_1=0.711$, $\delta_2=0.351$, $\nu_2=0.711$\\ (2.06), (2.06)\end{tabular} & \begin{tabular}[c]{@{}l@{}}$\delta=1.16$, $\nu=0.20$ \\ $\delta_1=0.464$, $\nu_1=0.540$, $\delta_2=0.464$, $\nu_2=0.540$ \\ (2.06), (2.06)\end{tabular}          & \begin{tabular}[c]{@{}l@{}}$\delta=1.16$, $\nu=0.20$\\ $\delta_1=0.351$, $\nu_1=0.711$, $\delta_2=0.351$, $\nu_2=0.711$\\ (2.06), (2.06)\end{tabular}            & \begin{tabular}[c]{@{}l@{}}$\delta=1.16$, $\nu=0.20$\\ $\delta_1=0.351$, $\nu_1=0.711$, $\delta_2=0.351$, $\nu_2=0.711$\\ (2.06), (2.06)\end{tabular} & \begin{tabular}[c]{@{}l@{}}$\delta=0.83$, $\nu=0.30$\\ (1.80)\end{tabular}   & $\delta=0.80$, $\nu=0.28$                                                                      \\ \cline{2-8} 
				& \begin{tabular}[c]{@{}l@{}}$\delta=0.9$, $\nu=0.27$\\ (1.76)\end{tabular}  & \begin{tabular}[c]{@{}l@{}}$\delta=1.47$, $\nu=0.17$\\ $\delta_1=0.37$, $\nu_1=0.66$, $\delta_2=0.37$, $\nu_2=0.66$\\ (2.48), (2.48)\end{tabular}     & \begin{tabular}[c]{@{}l@{}}$\delta=1.47$, $\nu=0.17$\\ $\delta_1=0.37$, $\nu_1=0.66$, $\delta_2=0.37$, $\nu_2=0.66$\\ (2.48), (2.48)\end{tabular}                & \begin{tabular}[c]{@{}l@{}}$\delta=1.47$, $\nu=0.17$\\ $\delta_1=0.37$, $\nu_1=0.66$, $\delta_2=0.37$, $\nu_2=0.66$\\ (2.48), (2.48)\end{tabular}                & \begin{tabular}[c]{@{}l@{}}$\delta=1.47$, $\nu=0.17$\\ $\delta_1=0.37$, $\nu_1=0.66$, $\delta_2=0.37$, $\nu_2=0.66$\\ (2.48), (2.48)\end{tabular}     & \begin{tabular}[c]{@{}l@{}}$\delta=0.92$, $\nu=0.27$\\ (1.01)\end{tabular}   & $\delta=0.91$, $\nu=0.26$                                                                      \\ \cline{2-8} 
				& N/A                                                                        & N/A                                                                                                                                                   & N/A                                                                                                                                                              & N/A                                                                                                                                                              & N/A                                                                                                                                                   & N/A                                                                          & \begin{tabular}[c]{@{}l@{}}None\\ (Base Training)\end{tabular}                                 \\ \cline{2-8} 
				& N/A                                                                        & N/A                                                                                                                                                   & N/A                                                                                                                                                              & N/A                                                                                                                                                              & N/A                                                                                                                                                   & N/A                                                                          & \begin{tabular}[c]{@{}l@{}}None\\ Base Network with Freb., \\ $\ell_2$ reg., $\lambda=0.05$\end{tabular} \\ \hline
			\end{tabular}
		}
		\label{tab:my-table10}
	\end{table}
\end{sidewaystable} 
\newpage
\subsection{The Lyapunov design parameters for the ResNet50 architecture (trained on the ImageNet dataset)} \label{app_13}
\begin{sidewaystable} 
	\begin{table}[H]
			\centering
		\caption{The hyper-parameters for the ResNet50 architecture (conv. layer, 4 blocks of sizes $[3, 4, 6, 3]$, linear layer) used in the experiments 
		\newline}
		\scalebox{.55}{
			\begin{tabular}{|l|l|l|l|l|l|l|l|}
				\hline
				ResNet          & Layer 1 (conv.)           & Block 1          & Block 2                & Block 3          & \multicolumn{1}{c|}{Block 4}     & Layer 5 (linear)          & \begin{tabular}[c]{@{}l@{}}Global Lyapunov\\ Property\end{tabular}       \\ \hline
				\multirow{5}{*}{\begin{tabular}[c]{@{}l@{}} Design\\ Parameters \end{tabular}   } & $\delta$, $\nu$    & \begin{tabular}[c]{@{}l@{}}$\delta$, $\nu$\\ $\delta_l$, $\nu_l$, $l=1,...,3$  \\ $3\times$(spect. norm reg.)\end{tabular}         & \begin{tabular}[c]{@{}l@{}} $\delta$, $\nu$, \\$\delta_l$, $\nu_l$, $l=1,...,4$\\ $4\times$(spect. norm reg.) \end{tabular}  & \begin{tabular}[c]{@{}l@{}}  $\delta$, $\nu$,\\ $\delta_l$, $\nu_l$, $l=1,...,6$\\ $6\times$(spect. norm reg.)  \end{tabular} & \begin{tabular}[c]{@{}l@{}}$\delta$, $\nu$\\ $\delta_l$, $\nu_l$, $l=1,...,3$ \\ $3\times$(spect. norm reg.)\end{tabular}          & \begin{tabular}[c]{@{}l@{}}$\delta$, $\nu$\\ (spect. norm reg.)\end{tabular} & $\delta$, $\nu$                                                                                \\ \cline{2-8} 
				
				& \begin{tabular}[c]{@{}l@{}}$\delta=0.87$, $\nu=0.29$\\ (1.60)\end{tabular} & \begin{tabular}[c]{@{}l@{}}$\delta=1.16$, $\nu=0.20$\\ $\delta_l=0.351$, $\nu_l=0.711$, $l=1,...,3$\\ $3\times(1.61)$ \end{tabular} & \begin{tabular}[c]{@{}l@{}}$\delta=1.16$, $\nu=0.20$\\ $\delta_l=0.351$, $\nu_l=0.711$, $l=1,...,4$\\ $4\times(1.61)$\end{tabular}          & \begin{tabular}[c]{@{}l@{}}$\delta=1.16$, $\nu=0.20$\\ $\delta_l=0.351$, $\nu_l=0.711$, $l=1,...,3$\\ $6\times(1.61)$\end{tabular}            & \begin{tabular}[c]{@{}l@{}}$\delta=1.16$, $\nu=0.20$\\ $\delta_l=0.351$, $\nu_l=0.711$, $l=1,...,3$\\ $3\times(1.61)$\end{tabular} & \begin{tabular}[c]{@{}l@{}}$\delta=0.88$, $\nu=0.28$\\ (1.60)\end{tabular}   & $\delta=0.89$, $\nu=0.28$                                                                       \\ \cline{2-8}

				& \begin{tabular}[c]{@{}l@{}}$\delta=0.93$, $\nu=0.27$\\ (2.0)\end{tabular}  & \begin{tabular}[c]{@{}l@{}}$\delta=1.16$, $\nu=0.20$\\ $\delta_l=0.351$, $\nu_l=0.711$, $l=1,...,3$\\ $3\times(2.06)$\end{tabular}     & \begin{tabular}[c]{@{}l@{}}$\delta=1.16$, $\nu=0.20$\\ $\delta_l=0.351$, $\nu_l=0.711$, $l=1,...,4$\\ $4\times(2.06)$\end{tabular}                & \begin{tabular}[c]{@{}l@{}}$\delta=1.16$, $\nu=0.20$\\ $\delta_l=0.351$, $\nu_l=0.711$, $l=1,...,6$\\ $6\times(2.06)$\end{tabular}                & \begin{tabular}[c]{@{}l@{}}$\delta=1.16$, $\nu=0.20$\\ $\delta_l=0.351$, $\nu_l=0.711$, $l=1,...,3$\\ $3\times(2.06)$\end{tabular}     & \begin{tabular}[c]{@{}l@{}}$\delta=0.94$, $\nu=0.26$ \\ (2.0)\end{tabular}   & $\delta=0.95$, $\nu=0.26$                                                                               \\ \cline{2-8} 
				
				& \begin{tabular}[c]{@{}l@{}}$\delta=0.99$, $\nu=0.25$\\ (2.52)\end{tabular}  & \begin{tabular}[c]{@{}l@{}}$\delta=1.47$, $\nu=0.17$\\ $\delta_l=0.37$, $\nu_l=0.66$, $l=1,...,3$\\ $3\times(2.48)$\end{tabular} & \begin{tabular}[c]{@{}l@{}}$\delta=1.47$, $\nu=0.17$\\ $\delta_l=0.37$, $\nu_l=0.66$, $l=1,...,4$\\ $4\times(2.48)$\end{tabular}               &\begin{tabular}[c]{@{}l@{}}$\delta=1.47$, $\nu=0.17$\\ $\delta_l=0.37$, $\nu_l=0.66$, $l=1,...,6$\\ $6\times(2.48)$\end{tabular}               &\begin{tabular}[c]{@{}l@{}}$\delta=1.47$, $\nu=0.17$\\ $\delta_l=0.37$, $\nu_l=0.66$, $l=1,...,3$\\ $3\times(2.48)$\end{tabular}     & \begin{tabular}[c]{@{}l@{}}$\delta=0.99$, $\nu=0.24$\\ (2.52)\end{tabular}   & $\delta=1.0$, $\nu=0.24$                \\ \cline{2-8} 
								
				& N/A                                                                        & N/A                                                                                                                                                   & N/A                                                                                                                                                              & N/A                                                                                                                                                              & N/A                                                                                                                                                   & N/A                                                                          & \begin{tabular}[c]{@{}l@{}}None\\ (Base Training)\end{tabular}                                 \\ \cline{2-8} 
				& N/A                                                                        & N/A                                                                                                                                                   & N/A                                                                                                                                                              & N/A                                                                                                                                                              & N/A                                                                                                                                                   & N/A                                                                          & \begin{tabular}[c]{@{}l@{}}None\\ Base Network with Freb., \\ $\ell_2$ reg., $\lambda=0.01$\end{tabular} \\ \hline
			\end{tabular}
		}
		\label{tab:my-table10}
	\end{table}
\end{sidewaystable} 
\newpage

\section{Experiment  results} \label{sec:experiment_results}
This appendix includes the experiment results for all the above architectures against FGM and Iterative PGD attacks. A pair of results are presented per DNN for the cases where a DNN was trained with or without adversarial training. 
\subsection{The experiment results for the Forward-Net architecture (trained on the MNIST dataset)}  \label{app_9}
\begin{table}[H]
		\centering
		\caption{Experiment results for the FGM attack \newline}
			\scalebox{0.84}{
	\begin{tabular}{|l|l|l|l|l|l|}
		\hline
		\multirow{2}{*}{Network Type}                                                                                                  & \multirow{2}{*}{Type of Attack} & \multirow{2}{*}{Type of Training}                                        & \multicolumn{3}{l|}{\begin{tabular}[c]{@{}l@{}}Accuracy on the Adversarial Test Dataset\\           (Attack Strength $\epsilon$)\end{tabular}} \\ \cline{4-6} 
		&                                 &                                                                          & $\epsilon=0.1$                                   & $\epsilon=0.2$                               & $\epsilon=0.3$                               \\ \hline
		$\delta=0.89$, $\nu=0.28$                                                                                                      & FGM                            &                                                                          & \textbf{0.966}                     &\textbf{0.953}                 & 0.934                                       \\ \hline
		$\delta=0.95$, $\nu=0.26$                                                                                                      & FGM                            &                                                                          & 0.962                                           & 0.951                                       & \textbf{0.935}                 \\ \hline
		$\delta=1.0$, $\nu=0.24$                                                                                                       & FGM                            &                                                                          & 0.948                                           & 0.939                                       & 0.927                                       \\ \hline
		\begin{tabular}[c]{@{}l@{}}Base Network with \\ Freb. $\ell_2$ reg., $\lambda=0.01$\end{tabular}                                  & FGM                            &                                                                          & 0.927                                           & 0.915                                       & 0.902                                       \\ \hline
		\begin{tabular}[c]{@{}l@{}}Base Training with \\ Freb. $\ell_2$ reg., $\lambda=0.05$\end{tabular}                                 & FGM                            &                                                                          & 0.870                                             & 0.859                                       & 0.840                                         \\ \hline
		\begin{tabular}[c]{@{}l@{}}Base Training \\ with Freb. $\ell_2$ reg., $\lambda=0.1$\end{tabular}                                  & FGM                            &                                                                          & 0.816                                            & 0.809                                       & 0.797                                       \\ \hline
		$\delta=0.89$, $\nu=0.28$                                                                                                      & FGM                            & \begin{tabular}[c]{@{}l@{}}with $\ell_2$ FGM  \\ Adv. Training\end{tabular} & \textbf{0.971}                 & \textbf{0.965}                 & \textbf{0.954}                 \\ \hline
		$\delta=0.95$, $\nu=0.26$                                                                                                      & FGM                            & \begin{tabular}[c]{@{}l@{}}with $\ell_2$ FGM \\ Adv. Training\end{tabular}  & 0.966                                           & 0.956                                       & 0.950                                       \\ \hline
		$\delta=1.0$, $\nu=0.24$                                                                                                       & FGM                            & \begin{tabular}[c]{@{}l@{}}with $\ell_2$ FGM  \\ Adv. Training\end{tabular} & 0.952                                           & 0.941                                       & 0.935                                       \\ \hline
		\begin{tabular}[c]{@{}l@{}}Base Network with \\ Freb. $\ell_2$ reg., $\lambda=0.01$\end{tabular} & FGM                            & \begin{tabular}[c]{@{}l@{}}with $\ell_2$ FGM \\ Adv. Training\end{tabular} & 0.908                                           & 0.898                                       & 0.885                                       \\ \hline
		\begin{tabular}[c]{@{}l@{}}Base Network with \\ Freb. $\ell_2$ reg., $\lambda=0.05$\end{tabular} & FGM                            & \begin{tabular}[c]{@{}l@{}}with $\ell_2$ FGM \\ Adv. Training\end{tabular} & 0.822                                           & 0.805                                       & 0.793                                       \\ \hline
		\begin{tabular}[c]{@{}l@{}}Base Network with \\ Freb. $\ell_2$ reg., $\lambda=0.1$\end{tabular}  & FGM                            & \begin{tabular}[c]{@{}l@{}}with $\ell_2$ FGM  \\ Training\end{tabular}      & 0.592                                           & 0.593                                        & 0.576                                        \\ \hline
	\end{tabular}
}
	\label{tab:my-table3}
\end{table}

\begin{table}[H]
		\centering
		\caption{Experiment results for the Iterative PGD attack (k=100)\newline}
			\scalebox{0.84}{
	\begin{tabular}{|l|l|l|l|l|l|}
		\hline
		\multirow{2}{*}{Network Type}                                                                                                  & \multirow{2}{*}{Type of Attack} & \multirow{2}{*}{Type of Training}                                        & \multicolumn{3}{l|}{\begin{tabular}[c]{@{}l@{}}Accuracy on the Adversarial Test Dataset\\           (Attack Strength $\epsilon$)\end{tabular}} \\ \cline{4-6} 
		&                                 &                                                                          & $\epsilon=0.1$                                   & $\epsilon=0.2$                               & $\epsilon=0.3$                               \\ \hline
		$\delta=0.89$, $\nu=0.28$                                                                                                      & PGD                            &                                                                          & \textbf{0.966}                     &\textbf{0.953}                 & 0.933                                       \\ \hline
		$\delta=0.95$, $\nu=0.26$                                                                                                      & PGD                            &                                                                          & 0.962                                         & 0.951                                     & \textbf{0.934}                 \\ \hline
		$\delta=1.0$, $\nu=0.24$                                                                                                       & PGD                            &                                                                          & 0.948                                          & 0.939                                    & 0.926                                     \\ \hline
		\begin{tabular}[c]{@{}l@{}}Base Network with \\ Freb. $\ell_2$ reg., $\lambda=0.01$\end{tabular}                                  & PGD                            &                                                                          &0.927                                           & 0.916                                      & 0.901                                       \\ \hline
		\begin{tabular}[c]{@{}l@{}}Base Training with \\ Freb. $\ell_2$ reg., $\lambda=0.05$\end{tabular}                                 & PGD                            &                                                                          & 0.869                                 & 0.858                                     &0.839                                  \\ \hline
		\begin{tabular}[c]{@{}l@{}}Base Training \\ with Freb. $\ell_2$ reg., $\lambda=0.1$\end{tabular}                                  & PGD                            &                                                                          & 0.815                                     & 0.809                                     & 0.796                                  \\ \hline
		$\delta=0.89$, $\nu=0.28$                                                                                                      & PGD                            & \begin{tabular}[c]{@{}l@{}}with $\ell_2$ PGD  \\ Adv. Training\end{tabular} & \textbf{0.970}                 & \textbf{0.963}                 & \textbf{0.954}                 \\ \hline
		$\delta=0.95$, $\nu=0.26$                                                                                                      & PGD                            & \begin{tabular}[c]{@{}l@{}}with $\ell_2$ PGD \\ Adv. Training\end{tabular}  &0.962                                         & 0.956                                      & 0.946                                \\ \hline
		$\delta=1.0$, $\nu=0.24$                                                                                                       & PGD                            & \begin{tabular}[c]{@{}l@{}}with $\ell_2$ PGD  \\ Adv. Training\end{tabular} & 0.950                                      & 0.943                                      & 0.933                                   \\ \hline
		\begin{tabular}[c]{@{}l@{}}Base Network with \\ Freb. $\ell_2$ reg., $\lambda=0.01$\end{tabular} & PGD                            & \begin{tabular}[c]{@{}l@{}}with $\ell_2$ PGD \\ Adv. Training\end{tabular} & 0.639                                        & 0.633                                   & 0.625                                   \\ \hline
		\begin{tabular}[c]{@{}l@{}}Base Network with \\ Freb.  $\ell_2$ reg., $\lambda=0.05$\end{tabular} & PGD                            & \begin{tabular}[c]{@{}l@{}}with $\ell_2$ PGD \\ Adv. Training\end{tabular} &0.113                                         & 0.113                                      & 0.113                                \\ \hline
		\begin{tabular}[c]{@{}l@{}}Base Network with \\ Freb.  $\ell_2$ reg., $\lambda=0.1$\end{tabular}  & PGD                            & \begin{tabular}[c]{@{}l@{}}with $\ell_2$ PGD  \\ Adv. Training\end{tabular}      & 0.113                                          & 0.113                                      & 0.113                                       \\ \hline
	\end{tabular}
}
	\label{tab:my-table4}
\end{table}

\subsection{The experiment results for the AlexNet architecture (trained on the CIFAR-10 dataset)}  \label{app_10}
\begin{figure}[H]
	\begin{center}
		\centering
		\begin{tikzpicture}[]
		
		\begin{groupplot}[
		group style={
			group name=myplot, group size=2 by 1,
			vertical sep=1.75cm,
		},
		enlarge x limits=true,
		]
		\centering
		
		\nextgroupplot[
		height=5cm,
		width=.45\columnwidth,
		ylabel={\small Accuracy under Attack},
		legend style={nodes={scale=0.75, transform shape},at={(0.15,0.45)},anchor=west},
		legend columns=1,
		]

		\addplot+[] table [x=epsilon, y=d_086_v_29, ] {pgdm_cifar10_advtrain_alexnet.dat};
		\addplot+[] table [x=epsilon, y=d_090_v_27, ] {pgdm_cifar10_advtrain_alexnet.dat};
		\addplot+[] table [x=epsilon, y=d_107_v_22,] {pgdm_cifar10_advtrain_alexnet.dat};
		\addplot+[dashed] table [x=epsilon, y=no_reg, ] {pgdm_cifar10_advtrain_alexnet.dat};
		\addplot+[magenta,mark options={magenta}] table [x=epsilon, y=l2_reg, green] {pgdm_cifar10_advtrain_alexnet.dat};
		
		\legend{$\delta=0.85,\nu=0.29$ \\ 
			$\delta=0.90,\nu=0.27$\\
			$\delta=1.07,\nu=0.22$\\
			Baseline \\ 
			$\ell_2$ Reg\\
		}
	
		\nextgroupplot[
		height=5cm,
		width=.45\columnwidth,
		legend style={nodes={scale=0.80, transform shape}, at={(0.25,0.25)},anchor=west},
		legend columns=1,
		]

		\addplot+[] table [x=epsilon, y=d_086_v_29, ] {fgsm_cifar10_advtrain_alexnet.dat};
		\addplot+[] table [x=epsilon, y=d_090_v_27, ] {fgsm_cifar10_advtrain_alexnet.dat};
		\addplot+[] table [x=epsilon, y=d_107_v_22, ] {fgsm_cifar10_advtrain_alexnet.dat};
		\addplot+[dashed] table [x=epsilon, y=no_reg, ] {fgsm_cifar10_advtrain_alexnet.dat};
		\addplot+[magenta,mark options={magenta}] table [x=epsilon, y=l2_reg] {fgsm_cifar10_advtrain_alexnet.dat};

		\end{groupplot}
		
		\end{tikzpicture}
	\end{center}
	\caption{Accuracy of model under PGD attack with adversarial training (left) and under FGM attack with adversarial training (right) using AlexNet on CIFAR10. Plots share the same legend, x-axis indicates the power of the attack $\epsilon$. } \label{fig:cifar10_pgdm_fgsm_results_adv}
\end{figure}
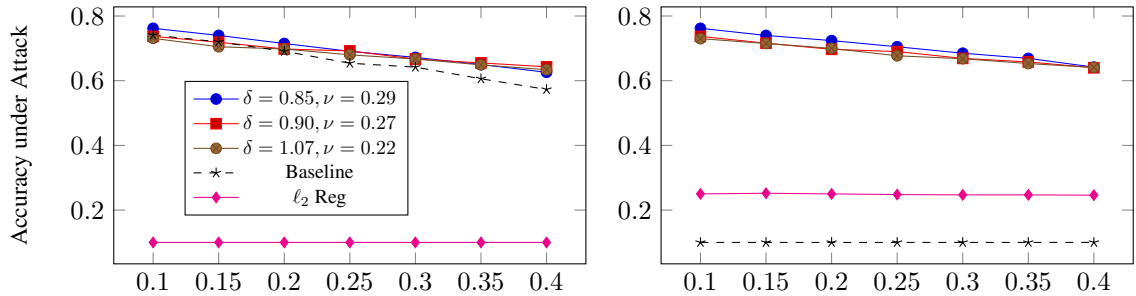

\begin{table}[H]
		\centering
		\caption{Experiment results for the FGM attack\newline}
	\scalebox{0.68}{
		\begin{tabular}{|l|l|l|l|l|l|l|l|l|l|}
			\hline
			\multirow{2}{*}{Network Type}                                                                        & \multirow{2}{*}{Type of Attack} & \multirow{2}{*}{Type of Training}                                        & \multicolumn{7}{l|}{\begin{tabular}[c]{@{}l@{}}Accuracy on the Adversarial Test Dataset\\           (Attack Strength $\epsilon$)\end{tabular}}                                                                              \\ \cline{4-10} 
			&                                 &                                                                          & $\epsilon=0.10$                  & $\epsilon=0.15$              & $\epsilon=0.20$              & $\epsilon=0.25$              & $\epsilon=0.30$              & $\epsilon=0.35$              & $\epsilon=0.40$               \\ \hline
			$\delta=0.85$, $\nu=0.29$                                                                            & FGM                    &                                                                          & \textbf{0.745}     &  \textbf{0.708}                     & 0.659                       & 0.612                     & 0.565                      & 0.515                       & 0.469                     \\ \hline
			$\delta=0.90$, $\nu=0.27$                                                                            & FGM                  &                                                                          & 0.733                         & 0.705 & \textbf{0.678} & \textbf{0.644} & \textbf{0.609} & 0.573                      & 0.536                      \\ \hline
			$\delta=1.07$, $\nu=0.22$                                                                          & FGM                    &                                                                          & 0.721                          & 0.698                     & 0.669                     & 0.639                      & 0.606                & \textbf{0.575} & \textbf{0.539} \\ \hline
			Base Network                                                                                         & FGM                   &                                                                          & 0.729                           & 0.680                  & 0.620                     & 0.563                     &0.508                   & 0.458                       & 0.424                  \\ \hline
			\begin{tabular}[c]{@{}l@{}}Base Training with \\ Freb. $\ell_2$ reg., $\lambda=0.01$\end{tabular}       & FGM                   &                                                                          &0.409                       & 0.407                   & 0.405              & 0.404                     & 0.397                      & 0.394                  & 0.391                  \\ \hline
			$\delta=0.85$, $\nu=0.29$                                                                            & FGM                    & \begin{tabular}[c]{@{}l@{}}with $\ell_2$ FGM \\ Adv. Training\end{tabular} &\textbf{0.762} & \textbf{0.740} & \textbf{0.724} & \textbf{0.705}                       & \textbf{0.685} &\textbf{0.669}                  & \textbf{0.642}                     \\ \hline
			$\delta=0.90$, $\nu=0.27$                                                                            & FGM                  & \begin{tabular}[c]{@{}l@{}}with $\ell_2$ FGM \\ Adv. Training\end{tabular} & 0.737                       & 0.716                     &0.697                     & 0.690 & 0.669                      & 0.658 & 0.640  \\ \hline
			$\delta=1.07$, $\nu=0.22$                                                                            & FGM                     & \begin{tabular}[c]{@{}l@{}}with $\ell_2$ FGM \\ Adv. Training\end{tabular} & 0.730                           & 0.715                    & 0.700                      & 0.677               & 0.668            &0.653                   & 0.641                       \\ \hline
			Base Network                                                                                         & FGM                   & \begin{tabular}[c]{@{}l@{}}with $\ell_2$ FGM \\ Adv. Training\end{tabular} & 0.100                         & 0.100          & 0.100                  & 0.100               & 0.100                & 0.100                   & 0.100  \\ \hline
			\begin{tabular}[c]{@{}l@{}}Base Training with \\ Freb. $\ell_2$ reg., $\lambda=0.01$\end{tabular}       & FGM                   & \begin{tabular}[c]{@{}l@{}}with $\ell_2$ FGM \\ Adv. Training\end{tabular} & 0.250                   &0.252           & 0.250                  & 0.248                          & 0.247         & 0.2478          & 0.246      \\ \hline
		\end{tabular}
	}
	\label{tab:my-table5}
\end{table}

\begin{table}[H]
		\centering
		\caption{Experiment results for the Iterative PGD attack (k=100) \newline}
	\scalebox{0.68}{
		\begin{tabular}{|l|l|l|l|l|l|l|l|l|l|}
			\hline
			\multirow{2}{*}{Network Type}                                                                        & \multirow{2}{*}{Type of Attack} & \multirow{2}{*}{Type of Training}                                        & \multicolumn{7}{l|}{\begin{tabular}[c]{@{}l@{}}Accuracy on the Adversarial Test Dataset\\           (Attack Strength $\epsilon$)\end{tabular}}                                                                              \\ \cline{4-10} 
			&                                 &                                                                          & $\epsilon=0.10$                  & $\epsilon=0.15$              & $\epsilon=0.20$              & $\epsilon=0.25$              & $\epsilon=0.30$              & $\epsilon=0.35$              & $\epsilon=0.40$               \\ \hline
			$\delta=0.85$, $\nu=0.29$                                                                            & PGD (k=100)                    &                                                                          & \textbf{0.743}     & 0.700                       & 0.645                        & 0.590                       & 0.530                       & 0.473                       & 0.415                        \\ \hline
			$\delta=0.90$, $\nu=0.27$                                                                            & PGD (k=100)                    &                                                                          & 0.731                           & \textbf{0.702} & \textbf{0.672} & \textbf{0.633} & \textbf{0.591} & 0.545                      & 0.504                        \\ \hline
			$\delta=1.07$, $\nu=0.22$                                                                          & PGD (k=100)                    &                                                                          & 0.720                           & 0.696                       & 0.662                       & 0.626                       & 0.590                       & \textbf{0.547} & \textbf{0.510} \\ \hline
			Base Network                                                                                         & PGD (k=100)                    &                                                                          & 0.730                            & 0.672                       & 0.608                       & 0.541                       & 0.478                       & 0.419                       & 0.367                        \\ \hline
			\begin{tabular}[c]{@{}l@{}}Base Training with \\ Freb. $\ell_2$ reg., $\lambda=0.01$\end{tabular}       & PGD (k=100)                    &                                                                          & 0.409                           & 0.407                       & 0.405                       & 0.402                       & 0.396                       & 0.393                       & 0.390                                   \\ \hline
			$\delta=0.85$, $\nu=0.29$                                                                            & PGD (k=100)                    & \begin{tabular}[c]{@{}l@{}}with $\ell_2$ PGD \\ Adv. Training\end{tabular} &\textbf{0.762} & \textbf{0.740} & \textbf{0.715} & 0.691                       & \textbf{0.672} & 0.650                       & 0.626                        \\ \hline
			$\delta=0.90$, $\nu=0.27$                                                                            & PGD (k=100)                    & \begin{tabular}[c]{@{}l@{}}with $\ell_2$ PGD \\ Adv. Training\end{tabular} & 0.737                           & 0.719                       & 0.698                       & \textbf{0.692} & 0.667                       & \textbf{0.655} & \textbf{0.643}  \\ \hline
			$\delta=1.07$, $\nu=0.22$                                                                           & PGD (k=100)                    & \begin{tabular}[c]{@{}l@{}}with $\ell_2$ PGD \\ Adv. Training\end{tabular} & 0.731                           & 0.705                       & 0.698                       & 0.680                       & 0.667                       & 0.649                       & 0.634                        \\ \hline
			Base Network                                                                                         & PGD (k=100)                    & \begin{tabular}[c]{@{}l@{}}with $\ell_2$ PGD \\ Adv. Training\end{tabular} & 0.741                           & 0.720                       & 0.691                       & 0.654                       & 0.642                       & 0.606                       & 0.573                       \\ \hline
			\begin{tabular}[c]{@{}l@{}}Base Training with \\ Freb. $\ell_2$ reg., $\lambda=0.01$\end{tabular}       & PGD (k=100)                    & \begin{tabular}[c]{@{}l@{}}with $\ell_2$ PGD \\ Adv. Training\end{tabular} & 0.100                              & 0.100                          & 0.100                          & 0.100                          & 0.100                          & 0.100                          & 0.100                           \\ \hline
		\end{tabular}
	}
	\label{tab:my-table6}
\end{table}

\subsection{The experiment results for the ResNet architecture (trained on the SVHN dataset)}  \label{app_12}
\begin{table}[H]
		\centering
		\caption{Experiment results for the Iterative PGD attack (k=100) \newline}
	\scalebox{0.65}{
		\begin{tabular}{|l|l|l|l|l|l|l|l|l|l|}
			\hline
			\multirow{2}{*}{Network Type}                                                                        & \multirow{2}{*}{Type of Attack} & \multirow{2}{*}{Type of Training}                                        & \multicolumn{7}{l|}{\begin{tabular}[c]{@{}l@{}}Accuracy on the Adversarial Test Dataset\\           (Attack Strength $\epsilon$)\end{tabular}}                                                                              \\ \cline{4-10} 
			&                                 &                                                                          & $\epsilon=0.10$                  & $\epsilon=0.15$              & $\epsilon=0.20$              & $\epsilon=0.25$              & $\epsilon=0.30$              & $\epsilon=0.35$              & $\epsilon=0.40$               \\ \hline
			$\delta=0.80$, $\nu=0.28$                                                                            & PGD (k=100)                          &                                                                          & \textbf{0.869}     &  \textbf{0.814}                     &\textbf{0.746}                      & \textbf{0.674}                  & \textbf{0.607}                    & \textbf{0.545}                      &0.486                    \\ \hline
			$\delta=0.91$, $\nu=0.26$                                                                            & PGD (k=100)                       &                                                                          & 0.862                   & 0.808 & 0.742 & \textbf{0.674} & 0.606 &0.544                    & \textbf{0.488}                      \\ \hline
			Base Network                                                                                         & PGD  (k=100)                         &                                                                          & 0.850                         & 0.783               & 0.711                   & 0.631                 &0.565                  & 0.493                       & 0.431                  \\ \hline
			\begin{tabular}[c]{@{}l@{}}Base Training with \\ Freb. $\ell_2$ reg., $\lambda=0.01$\end{tabular}       & PGD (k=100)                   &                                                                          &0.850                & 0.783            & 0.719        &0.640                  &  0.571                 & 0.509                  & 0.455          \\ \hline
		\end{tabular}
	}
	\label{tab:my-table9}
\end{table}

\section{Robustness performance against  the Carlini \& Wagner (C\&W)  attack}  \label{app_11}
We tried our approach against the C\&W attack. The parameters were chosen according to the tutorials presented in \cite{papernot2018cleverhans}. This appendix includes the results for this experiment.
\begin{table}[H]
		\centering
		\caption{C\&W attack parameters: \{learning rate: 0.001, initial const: 0.01, max iter: 500\} \newline}
	\begin{tabular}{|l|l|l|l|}
		\hline
		Network Type/Dataset                                                                                & \multirow{2}{*}{\begin{tabular}[c]{@{}l@{}}Accuracy on the \\ adversarial dataset\end{tabular}} & Network Type/Dataset                                                                           & \multirow{2}{*}{\begin{tabular}[c]{@{}l@{}}Accuracy on the \\ adversarial dataset\end{tabular}} \\ \cline{1-1} \cline{3-3}
		AlexNet/Cifar-10                                                                                    &                                                                                                 & ForwardNet/MNIST                                                                               &                                                                                                 \\ \hline
		$\delta=0.86$, $\nu=0.29$                                                                           & 0.302                                                                                           & $\delta=0.89$, $\nu=0.28$                                                                      & 0.375                                                                                           \\ \hline
		$\delta=0.90$, $\nu=0.27$                                                                           & \textbf{0.335}                                                                                           & $\delta=0.95$, $\nu=0.26$                                                                      & 0.466                                                                                          \\ \hline
		$\delta=1.07$, $\nu=0.22$                                                                          & 0.332                                                                                           & $\delta=1.0$, $\nu=0.24$                                                                       & \textbf{0.470}                                                                                            \\ \hline
		Base Network                                                                                        & 0.265                                                                                           & Base Network                                                                                   & 0.382                                                                                           \\ \hline
		\begin{tabular}[c]{@{}l@{}}Base Training with\\ Freb. $\ell_2$ reg., $\lambda=0.01$\end{tabular}       & 0.157                                                                                           & \begin{tabular}[c]{@{}l@{}}Base Training with \\ Freb. $\ell_2$ reg., $\lambda=0.05$\end{tabular} & 0.318                                                                                         \\ \hline
		\begin{tabular}[c]{@{}l@{}}Base Training with \\ spec. reg. \textless 1 for all layers\end{tabular} & 0.281                                                                                          &                                                                                                &                                                                                                 \\ \hline
	\end{tabular}
	\label{tab:my-table7}
\end{table}

\begin{table}[H]
		\centering
		\caption{C\&W attack parameters: \{learning rate: 0.0001, initial const: 0, max iter: 500\} \newline}
	\begin{tabular}{|l|l|l|l|}
		\hline
		Network Type/Dataset                                                                                & \multirow{2}{*}{\begin{tabular}[c]{@{}l@{}}Accuracy on the \\ adversarial dataset\end{tabular}} & Network Type/Dataset                                                                           & \multirow{2}{*}{\begin{tabular}[c]{@{}l@{}}Accuracy on the \\ adversarial dataset\end{tabular}} \\ \cline{1-1} \cline{3-3}
		AlexNet/Cifar-10                                                                                    &                                                                                                 & ForwardNet/MNIST                                                                               &                                                                                                 \\ \hline
		$\delta=0.86$, $\nu=0.29$                                                                           & \textbf{0.787}                                                                                         & $\delta=0.89$, $\nu=0.28$                                                                      & 0.970                                                                                          \\ \hline
		$\delta=0.90$, $\nu=0.27$                                                                           & 0.755                                                                                          & $\delta=0.95$, $\nu=0.26$                                                                      & \textbf{0.975}                                                                                          \\ \hline
		$\delta=1.07$, $\nu=0.22$                                                                          & 0.740                                                                                          & $\delta=1.0$, $\nu=0.24$                                                                       & 0.951                                                                                          \\ \hline
		Base Network                                                                                        & 0.717                                                                                          & Base Network                                                                                   & 0.934                                                                                          \\ \hline
		\begin{tabular}[c]{@{}l@{}}Base Training with\\ Freb. $\ell_2$ reg., $\lambda=0.01$\end{tabular}       & 0.407                                                                                          & \begin{tabular}[c]{@{}l@{}}Base Training with \\ Freb. $\ell_2$ reg., $\lambda=0.05$\end{tabular} & 0.872                                                                                         \\ \hline
		\begin{tabular}[c]{@{}l@{}}Base Training with \\ spec. reg. \textless 1 for all layers\end{tabular} & 0.441                                                                                          &                                                                                                &                                                                                                 \\ \hline
	\end{tabular}
	\label{tab:my-table8}
\end{table}

\section{Robustness performance comparison with prior spectral normalization approaches }
\label{app_14}
This appendix details a set of experiments performed to provide a robustness comparison between our proposed approach and the works given in \cite{farnia2018generalizable}, \cite{qian2018l2}, \cite{yoshida2017spectral} and \cite{cisse2017parseval}.  \cite{qian2018l2}, \cite{farnia2018generalizable} propose training networks with the same spectral regularization enforced across the entire network where $\rho(W_l)\leq\beta$ for $l=1,...,n$ and $\beta$ is a constant. We select 3 values of $\beta$ from their papers: $\beta=1.0, 1.6,2.0$. The 2 works given in \cite{yoshida2017spectral} and \cite{cisse2017parseval} may be seen as subsets of the works given in \cite{qian2018l2} and \cite{farnia2018generalizable}, where  $\rho(W_l)\leq\beta$ for $l=1,...,n$ and $\beta=1.0$ are selected. All the aforementioned works are special cases of our proposed Lyapunov robust solution. Our experiments confirm that our approach provides 3 main benefits that the other works do not provide. 1- Our work provides the theory, reasoning and interpretation behind why spectral regularization enhances robustness, and in particular how the spectral regularization hyper-parameters for each layer should be selected. 2- Our work provides a higher level of flexibility and freedom in selecting the hyper-parameter for training based the interpretations and reasoning behind our theory and work. 3- Our proposed approach produces trained DNNs that are more robust and perform better than the other works in this area.
\begin{table}[H]
		\centering
		\caption{The hyper-parameters for the AlexNet architecture used in the experiments \newline}
	\scalebox{0.6}{
		\begin{tabular}{|l|l|l|l|l|l|l|}
			\hline
			AlexNet:                           & Layer 1 (conv.)                                                             & Layer 2 (conv.)                                                             & Layer 3 (conv.)                                                             & Layer 4 (linear)                                                            & Layer 5 (linear)                                                            & \begin{tabular}[c]{@{}l@{}}Global Lyapunov\\ Property\end{tabular}                                           \\ \hline
			\multirow{7}{*}{\begin{tabular}[c]{@{}l@{}} Design\\ Parameters \end{tabular}   } & \begin{tabular}[c]{@{}l@{}}$\delta$, $\nu$ \\ (spect. norm reg.)\end{tabular} & \begin{tabular}[c]{@{}l@{}}$\delta$, $\nu$\\  (spect. norm reg.)\end{tabular} & \begin{tabular}[c]{@{}l@{}}$\delta$, $\nu$\\  (spect. norm reg.)\end{tabular} & \begin{tabular}[c]{@{}l@{}}$\delta$, $\nu$ \\ (spect. norm reg.)\end{tabular} & \begin{tabular}[c]{@{}l@{}}$\delta$, $\nu$ \\ (spect. norm reg.)\end{tabular} & $\delta$, $\nu$                                                                 \\ \cline{2-7} 
			& \begin{tabular}[c]{@{}l@{}}$\delta=0.81$, $\nu=0.30$\\ (1.49)\end{tabular}      & \begin{tabular}[c]{@{}l@{}}$\delta=0.96$, $\nu=0.20$\\ (1.49)\end{tabular}      & \begin{tabular}[c]{@{}l@{}}$\delta=0.96$, $\nu=0.20$\\ (1.49)\end{tabular}      & \begin{tabular}[c]{@{}l@{}}$\delta=0.70$, $\nu=0.35$\\ (2.12)\end{tabular}      & \begin{tabular}[c]{@{}l@{}}$\delta=0.80$, $\nu=0.23$\\ (3.14)\end{tabular}      & \begin{tabular}[c]{@{}l@{}}$\delta=0.79$, $\nu=0.29$\end{tabular}                 \\ \cline{2-7} 
			& \begin{tabular}[c]{@{}l@{}}$\delta=0.86
				$, $\nu=0.29$\\ (1.59)\end{tabular}      & \begin{tabular}[c]{@{}l@{}}$\delta=0.86
				$, $\nu=0.29$\\ (1.89)\end{tabular}      & \begin{tabular}[c]{@{}l@{}}$\delta=0.96
				$, $\nu=0.26$\\ (1.62)\end{tabular}      & \begin{tabular}[c]{@{}l@{}}$\delta=0.96$, $\nu=0.26$\\ (1.62)\end{tabular}      & \begin{tabular}[c]{@{}l@{}}$\delta=0.90$, $\nu=0.27$\\ (2.81)\end{tabular}      & \begin{tabular}[c]{@{}l@{}}$\delta=0.89$, $\nu=0.28$\end{tabular}                                                                             \\ \cline{2-7} 
			& \begin{tabular}[c]{@{}l@{}}$\delta=0.88
				$, $\nu=0.28$\\ (1.59)\end{tabular}      & \begin{tabular}[c]{@{}l@{}}$\delta=0.89
				$, $\nu=0.28$\\ (1.89)\end{tabular}      & \begin{tabular}[c]{@{}l@{}}$\delta=0.76
				$, $\nu=0.32$\\ (2.52)\end{tabular}      & \begin{tabular}[c]{@{}l@{}}$\delta=0.96
				$, $\nu=0.26$\\ (1.62)\end{tabular}      & \begin{tabular}[c]{@{}l@{}}$\delta=0.91$, $\nu=0.27$\\ (1.80)\end{tabular}      & \begin{tabular}[c]{@{}l@{}}$\delta=0.90$, $\nu=0.27$\end{tabular}                                     \\ \cline{2-7} 
			& \begin{tabular}[c]{@{}l@{}}N/A, N/A\\ (1.0)\end{tabular}                    & \begin{tabular}[c]{@{}l@{}}N/A, N/A\\ (1.0)\end{tabular}                    & \begin{tabular}[c]{@{}l@{}}N/A, N/A\\ (1.0)\end{tabular}                    & \begin{tabular}[c]{@{}l@{}}N/A, N/A\\ (1.0)\end{tabular}                    & \begin{tabular}[c]{@{}l@{}}N/A, N/A\\ (1.0)\end{tabular}                    & \begin{tabular}[c]{@{}l@{}}None\\ (Training with spectral\\ reg. \textless{}1.0 across all layers)\end{tabular}                                          \\ \cline{2-7} 
			& \begin{tabular}[c]{@{}l@{}}N/A, N/A\\ (1.6)\end{tabular}                                                            &  \begin{tabular}[c]{@{}l@{}}N/A, N/A\\ (1.6)\end{tabular}                                                                             &  \begin{tabular}[c]{@{}l@{}}N/A, N/A\\ (1.6)\end{tabular}                                                                     &  \begin{tabular}[c]{@{}l@{}}N/A, N/A\\ (1.6)\end{tabular}                                                                   &  \begin{tabular}[c]{@{}l@{}}N/A, N/A\\ (1.6)\end{tabular}                                                                &  \begin{tabular}[c]{@{}l@{}}None\\ (Training with spectral\\ reg. \textless{}1.6 across all layers)\end{tabular}       
			 \\ \cline{2-7} 
			& \begin{tabular}[c]{@{}l@{}}N/A, N/A\\ (2.0)\end{tabular}                                                            &  \begin{tabular}[c]{@{}l@{}}N/A, N/A\\ (2.0)\end{tabular}                                                                             &  \begin{tabular}[c]{@{}l@{}}N/A, N/A\\ (2.0)\end{tabular}                                                                     &  \begin{tabular}[c]{@{}l@{}}N/A, N/A\\ (2.0)\end{tabular}                                                                   &  \begin{tabular}[c]{@{}l@{}}N/A, N/A\\ (2.0)\end{tabular}                                                                &  \begin{tabular}[c]{@{}l@{}}None\\ (Training with spectral\\ reg. \textless{}2.0 across all layers)\end{tabular}    
			      \\ \hline
		\end{tabular}
	}
	\label{tab:my-table11}
\end{table}

\begin{table}[H]
		\centering
		\caption{Experiment results for the FGM attack \newline}
	\scalebox{0.71}{
		\begin{tabular}{|l|l|l|l|l|l|l|l|l|}
			\hline
			\multirow{2}{*}{Network Type}                                                                        & \multirow{2}{*}{Type of Attack} & \multicolumn{7}{l|}{\begin{tabular}[c]{@{}l@{}}Accuracy on the Adversarial Test Dataset\\           (Attack Strength $\epsilon$)\end{tabular}}                                                                              \\ \cline{3-9} 
			&              &                                                                                       $\epsilon=0.10$                  & $\epsilon=0.15$              & $\epsilon=0.20$              & $\epsilon=0.25$              & $\epsilon=0.30$              & $\epsilon=0.35$              & $\epsilon=0.40$               \\ \hline
			$\delta=0.79$, $\nu=0.29$                                                                & FGM                    &                                                            0.720&\textbf{0.704}      &0.678               & 0.646                 & \textbf{0.614}               & 0.579     &0.547 \\ \hline
			$\delta=0.89$, $\nu=0.28$                                                                    & FGM                    &                                                                         \textbf{0.722}              &0.701 & 0.676 & \textbf{0.647} & 0.613 & \textbf{0.583}                  & \textbf{0.550}    \\ \hline
			$\delta=0.90$, $\nu=0.27$& FGM                   &                                                                           0.721                    & 0.703                    &\textbf{ 0.703}                & 0.643                  & 0.609                 & 0.578& 0.548         \\ \hline
			\begin{tabular}[c]{@{}l@{}}Base Training with \\ spec. reg. \textless{}1 across all layers\end{tabular} & FGM                 &                                                                           0.437                           & 0.436                       & 0.432                       & 0.430                         & 0.427                       & 0.425                       & 0.419                         \\ \hline
			\begin{tabular}[c]{@{}l@{}}Base Training with \\ spec. reg. \textless{}1.6 across all layers\end{tabular}       & FGM                   &                                                                       0.665          & 0.656               & 0.635               & 0.614           & 0.589            & 0.562                 & 0.538  \\ \hline
			\begin{tabular}[c]{@{}l@{}}Base Training with \\ spec. reg. \textless{}2.0 across all layers\end{tabular}       & FGM                   &                                                                   0.719      & 0.701   & 0.673             & 0.641     & 0.608         &0.569            &0.534      \\ \hline
		\end{tabular}
	}
	\label{tab:my-table12}
\end{table}

\begin{table}[H]
		\centering
		\caption{Experiment results for the Iterative PGD attack (k=100) \newline}
	\scalebox{0.71}{
	\begin{tabular}{|l|l|l|l|l|l|l|l|l|}
		\hline
		\multirow{2}{*}{Network Type}                                                                        & \multirow{2}{*}{Type of Attack} & \multicolumn{7}{l|}{\begin{tabular}[c]{@{}l@{}}Accuracy on the Adversarial Test Dataset\\           (Attack Strength $\epsilon$)\end{tabular}}                                                                              \\ \cline{3-9} 
		&              &                                                                                       $\epsilon=0.10$                  & $\epsilon=0.15$              & $\epsilon=0.20$              & $\epsilon=0.25$              & $\epsilon=0.30$              & $\epsilon=0.35$              & $\epsilon=0.40$               \\ \hline
		$\delta=0.79$, $\nu=0.29$                                                                & PGD                    &                                                            0.720 &\textbf{0.703}         & \textbf{0.671}         & 0.635               &\textbf{0.599}                  & 0.556         & 0.520            \\ \hline
		$\delta=0.89$, $\nu=0.28$                                                                    & PGD                    &                                                                         \textbf{0.722}              & 0.698 & 0.670 & \textbf{0.638} & 0.598 & \textbf{0.563}                  & \textbf{0.523}    \\ \hline
		$\delta=0.90$, $\nu=0.27$& PGD                   &                                                                      0.721                  & 0.701                  & 0.670              &0.633        & 0.593              & 0.557 & 0.522            \\ \hline
		\begin{tabular}[c]{@{}l@{}}Base Training with \\ spec. reg. \textless{}1 across all layers\end{tabular} & PGD                 &                                                                           0.437                           & 0.436                       & 0.432                       & 0.430                         & 0.427                       & 0.425                      & 0.419                         \\ \hline
		\begin{tabular}[c]{@{}l@{}}Base Training with \\ spec. reg. \textless{}1.6 across all layers\end{tabular}       & PGD                   &                                                  0.665                                  &         0.654             & 0.633               & 0.611         &        0.580  &  0.554 &0.522  \\ \hline
		\begin{tabular}[c]{@{}l@{}}Base Training with \\ spec. reg. \textless{}2.0 across all layers\end{tabular}       & PGD                   &                                                                       0.717    & 0.696              & 0.667             & 0.629   & 0.587        & 0.543                & 0.500     \\ \hline
	\end{tabular}
}
	\label{tab:my-table13}
\end{table}

\begin{table}[H]
		\centering
		\caption{The test accuracy of the DNNs after training \newline}
			\scalebox{0.83}{
	\begin{tabular}{|l|l|l|l|}
		\hline
		Network Type & \begin{tabular}[c]{@{}l@{}}Test Accuracy \\ on the Clean Dataset\end{tabular} & Network Type & \begin{tabular}[c]{@{}l@{}}Test Accuracy\\  on the Clean Dataset\end{tabular} \\ \hline
		$\delta=0.79$, $\nu=0.29$  & 0.74                                                                          &       	\begin{tabular}[c]{@{}l@{}}Base Training with \\ spec. reg. \textless{}1 across all layers\end{tabular}        & 0.44                                                                          \\ \hline
	 $\delta=0.89$, $\nu=0.28$	& 0.74                                                                          &     	\begin{tabular}[c]{@{}l@{}}Base Training with \\ spec. reg. \textless{}1.6 across all layers\end{tabular}          & 0.68                                                                          \\ \hline
		$\delta=0.90$, $\nu=0.27$ & 0.74                                                                          &            \begin{tabular}[c]{@{}l@{}}Base Training with \\ spec. reg. \textless{}2.0 across all layers\end{tabular}   & 0.74                                                                          \\ \hline
	\end{tabular}
	\label{tab:my-table14}
}
\end{table}

\section{Further details on the experiments} \label{app_15} \label{sec:experiment_details}
We validate our results by training several 3-layer fully connected forward-nets on the MNIST data set using the Adam optimizer with a learning rate of $0.001$. Further, we train several AlexNet architectures on the CIFAR10 data set using Stochastic Gradient Descent optimization with a learning rate of $0.01$ and momentum of $0.9$. Additionally, we train several ResNet architectures on the SVHN data set and ResNet50 architectures on the Imagenet data set using Stochastic Gradient Descent optimization with a learning rate of $0.01$ and momentum of $0.9$. All networks are trained for $200$ epochs. The pixel values for the input images are normalized to take values in $[-0.5,+0.5]$. When considering the baseline defense of weight decay (i.e., $\ell_2$ regularized weights) networks, we performed cross-validation to select $\lambda \in [0.01, 0.05, 0.1]$. The parameters for our robust Lyapunov approach for each network architecture used in the experiments are specified in Appendix \ref{sec:experiment_design_details}. Baseline represents a DNN with no regularization enforced during training. 

All experiments were performed on a single Nvidia Tesla V100-SXM2-16GB GPU. The MNIST data set was downloaded from \url{http://yann.lecun.com/exdb/mnist/}. The data is randomly divided into the training, testing and validation data sets of size: 60000, 10000 and 5000 receptively. The CIFAR10 data set was downloaded from \url{https://www.cs.toronto.edu/~kriz/cifar.html}. The data is randomly divided into the training, testing and validation data sets of size: 45000, 10000 and 5000 receptively. The SVHN data set was downloaded from \url{http://ufldl.stanford.edu/housenumbers/}. The data is randomly divided into the training, testing and validation data sets of size: 73257, 26032 and 500 receptively. The ImageNet data set was downloaded from \url{http://image-net.org/download}. No sample was excluded. 
\end{document}